\documentclass{article}

\usepackage{hyperref}       

\usepackage[accepted]{icml2023}

\usepackage[utf8]{inputenc} 
\usepackage[T1]{fontenc}    
\usepackage{url}            
\usepackage{booktabs}       
\usepackage{amsfonts}       
\usepackage{nicefrac}       
\usepackage{microtype}      
\usepackage{bbding}
\usepackage{multirow}

\usepackage{array}
\usepackage{amsmath}
\usepackage{amsfonts}
\usepackage{amsthm}
\usepackage{bbm}
\usepackage{syntax}

\usepackage[epsilon, altpo]{backnaur}

\usepackage[capitalize,noabbrev]{cleveref}

\usepackage{caption}
\usepackage{subcaption}
\usepackage{graphicx}
\usepackage{wrapfig}

\newcommand{\R}{\mathbb{R}}
\newcommand{\St}{\mathcal{S}}
\newcommand{\A}{\mathcal{A}}
\newcommand\Alpha{\mathrm{A}}
\newcommand{\True}{\top}
\newcommand{\False}{\bot}
\newcommand{\rlang}{RLang}
\newcommand{\E}{\mathbb{E}}

\definecolor{lightpink}{RGB}{252, 210, 230}
\definecolor{lightred}{RGB}{250, 203, 182}
\definecolor{lightblue}{RGB}{210, 228, 252}
\definecolor{lightyellow}{RGB}{255, 247, 189}
\definecolor{lightgreen}{RGB}{210, 252, 222}

\definecolor{darkpink}{RGB}{204, 0, 102}
\definecolor{darkred}{RGB}{204, 0, 0}
\definecolor{skyblue}{RGB}{0, 128, 255}
\definecolor{darkgreen}{RGB}{0, 153, 76}
\definecolor{purple}{RGB}{102, 0, 204}
\definecolor{teal}{RGB}{0, 153, 153}

\definecolor{backgroundcolor}{RGB}{245, 245, 245}

\usepackage{listings}

\lstdefinestyle{rlang}{
    backgroundcolor=\color{backgroundcolor},
    basicstyle=\ttfamily\small,
    breakatwhitespace=false,         
    breaklines=true,                 
    captionpos=b,                    
    keepspaces=true,                 
    numbers=none,                    
    numbersep=5pt,
    numberstyle=\tiny,
    showspaces=false,                
    showstringspaces=false,
    showtabs=false,                  
    tabsize=2,
    literate={\ \ }{{\ }}1,
    morecomment=[l]{####},
    xleftmargin=.2in,
    xrightmargin=.2in,
    belowskip=-0.35em,
    morekeywords={Factor, Markov, Feature, Constant, Proposition, if, else, elif, Policy, Effect, Option, init, until, and, or, not, S, S', A, Reward, Value, Goal, P, Dynamics, in, ActionRestriction, Class, Object}
}

\newcommand{\specialcell}[2][c]{%
  \begin{tabular}[#1]{@{}l@{}}#2\end{tabular}}

\icmltitlerunning{RLang: A Declarative Language for Expressing Partial World Models}

\begin{document}

\twocolumn[
\icmltitle{RLang: A Declarative Language for Describing \\ Partial World Knowledge to Reinforcement Learning Agents}



\icmlsetsymbol{equal}{*}

\begin{icmlauthorlist}
\icmlauthor{Rafael Rodriguez-Sanchez}{equal,brown}
\icmlauthor{Benjamin A. Spiegel}{equal,brown}
\icmlauthor{Jennifer Wang}{brown}\\
\icmlauthor{Roma Patel}{brown}
\icmlauthor{Stefanie Tellex}{brown}
\icmlauthor{George Konidaris}{brown}
\end{icmlauthorlist}

\icmlaffiliation{brown}{Department of Computer Science, Brown University, Providence, RI, USA}

\icmlcorrespondingauthor{Benjamin Spiegel}{bspiegel@cs.brown.edu}

\icmlkeywords{Language, Reinforcement Learning, MDP}

\vskip 0.3in
]

\printAffiliationsAndNotice{\icmlEqualContribution}
\begin{abstract}
We introduce RLang, a domain-specific language (DSL) for communicating  domain knowledge to an RL agent. Unlike  existing RL DSLs that ground to \textit{single} elements of a decision-making formalism (e.g., the reward function or policy), RLang can specify information about every element of a Markov decision process. 
We define precise syntax and grounding semantics for RLang, and provide a parser that grounds RLang programs to an algorithm-agnostic \textit{partial} world model and policy that can be exploited by an RL agent. 
We provide a series of example RLang programs demonstrating how different RL methods can exploit the resulting knowledge,
encompassing model-free and model-based tabular algorithms, policy gradient and value-based methods, hierarchical approaches, and deep methods.

\end{abstract}

\section{Introduction}
Reinforcement learning tasks are often impractically hard to solve \textit{tabula rasa}. Fortunately, even a small amount of prior knowledge about the world---knowledge that is often either seemingly obvious or easy for a human to produce---can dramatically improve learning. For instance, knowing about the action dynamics of a game (e.g., you can jump to avoid falling into pits) or properties of its state (e.g., that a particular state variable indicates a block of lava) can prevent an agent from making poor decisions. In long-horizon tasks, especially those with sparse rewards, such knowledge may even be a prerequisite for finding an acceptable policy.

Languages, both formal and natural, have been used in various ways to add prior knowledge into decision-making \cite{luketina2019survey}. 
Formal languages benefit from unambiguous syntax and semantics, and can therefore be reliably used to represent knowledge. These have proven useful in specifying advice to agents in the form of hints about actions \cite{maclin1996creating} or policy structure \cite{andreas2017modular, sun2020program}.
Communicating using natural language would be more intuitive, but that requires converting natural language sentences into grounded knowledge usable by the agent; most approaches in this area can ground limited subsets of natural language (e.g., only commands).
For example, for describing task objectives \cite{artzi2013weakly, patel20}, or other individual task components such as rewards \cite{goyal2019using, sumers2021learning} and policies \cite{branavan2010reading}.
All of the above approaches provide information about a single component of a chosen decision-making formalism; there exists no unified framework able to express information about \emph{all} the components of a task.  

We therefore introduce RLang, a domain-specific language (DSL) that can specify information about every component of a Markov decision process (MDP), including flat and hierarchical policies, state factors, state features, transition functions, and reward functions. \rlang{} is human-readable and compiles into simple data structures that can be accessed by any learning algorithm. We explain how to write statements that inform each MDP component, and  demonstrate \rlang's versatility through a series of example programs expressing different types of domain knowledge. We also show how to integrate this knowledge into several RL methods to improve learning performance.\footnote{RLang source code, documentation, and examples are available at \url{rlang.ai}.}


\begin{figure*}[t]
    \centering
    \begin{subfigure}[b]{0.4\textwidth}
         \includegraphics[width=\linewidth]{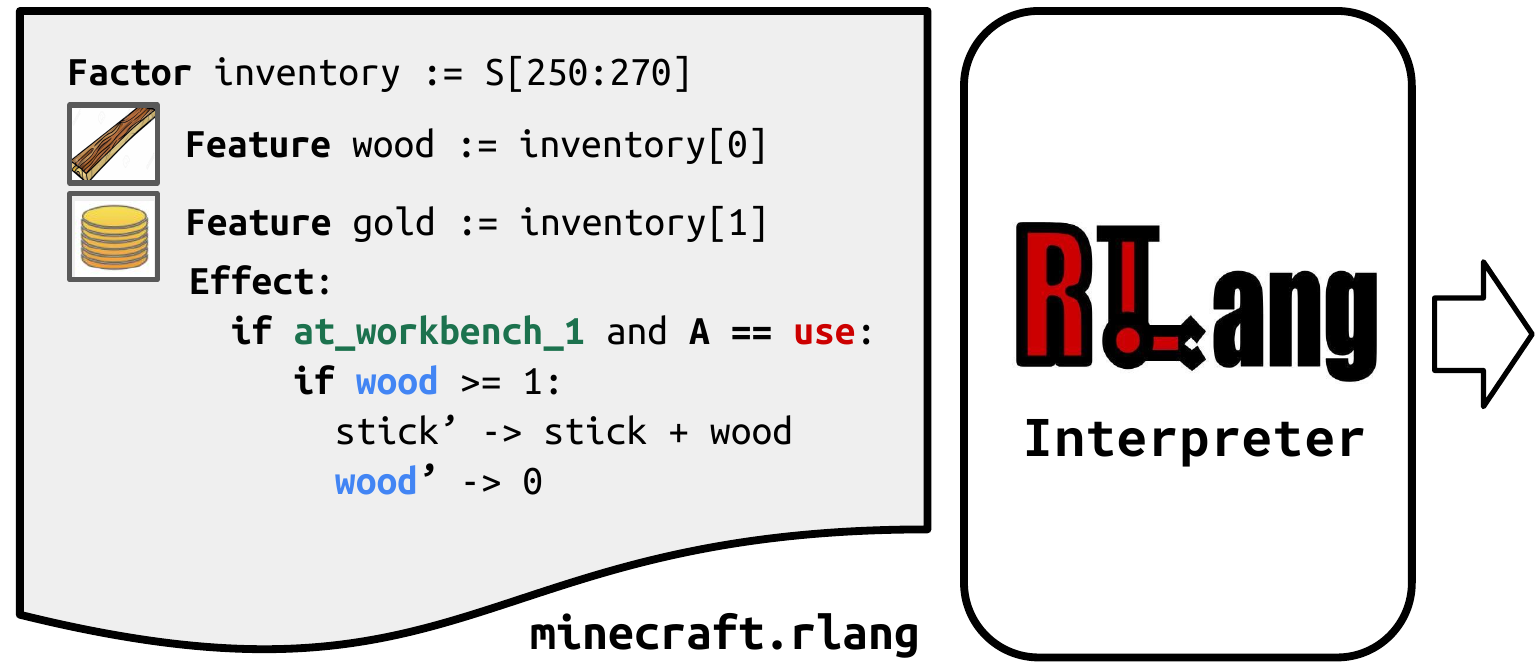}
    \end{subfigure}
    \begin{subfigure}[b]{0.58\textwidth}
         \includegraphics[width=\linewidth]{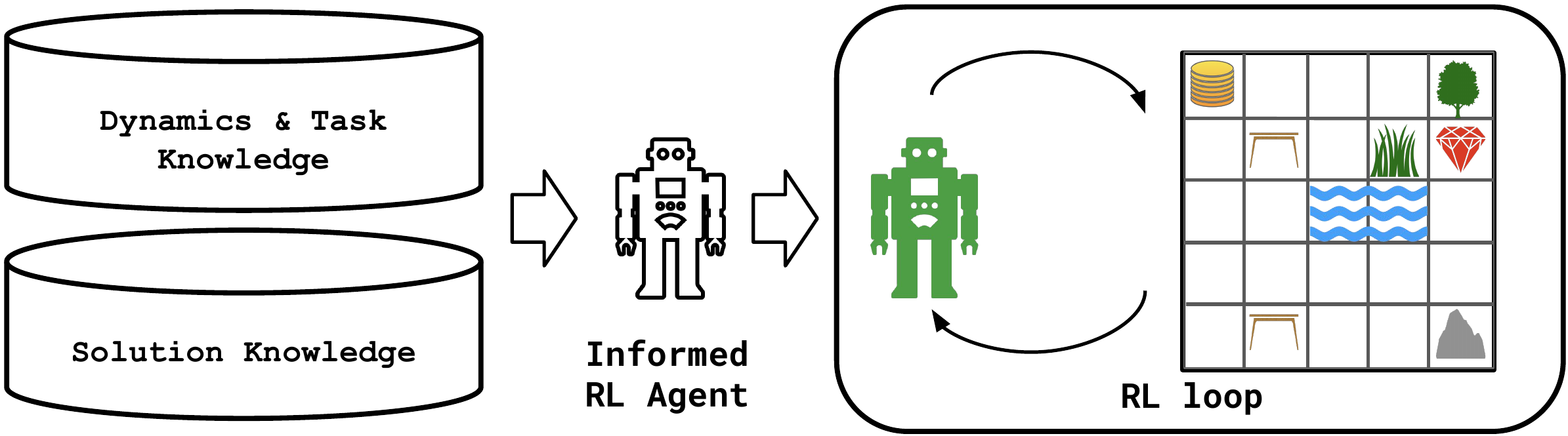}
    \end{subfigure}
    \caption{RLang provides users with a precise language to provide domain knowledge to RL agents. RLang programs are parsed by the interpreter to create a partial model of the MDP (\texttt{Dynamics and Task Knowledge}) and its solution (\texttt{Solution Knowledge}). Finally, the knowledge objects inform an RL agent that can leverage the grounded information during its learning loop.}
    \label{fig:main}
\end{figure*}

\section{Background}
\textbf{Domain-Specific Languages} Domain-specific languages (DSLs) are formal languages designed to specify information relevant to a target domain. Compared to general-purpose programming languages like Python \cite{van1995python} and C \cite{kernigham1973c}, DSLs typically contain a smaller set of narrower semantics better suited to a specific application. That is, DSLs sacrifice computational expressivity for ease-of-use within a particular domain.  Commonly-used DSLs include the Standard Query Language (SQL) used for querying relational databases and the Planning Domain Definition Language (PDDL; \citealt{ghallab1998pddl}) for defining  planning tasks.

\textbf{Decision-Making Formalisms} Reinforcement learning tasks are typically modeled as Markov Decision Processes (MDPs; \citealp{puterman1990markov}),  defined by a tuple $(\mathcal{S}, \mathcal{A}, R, T, \gamma)$, where $\mathcal{S}$ is a set of states, $\mathcal{A}$ is a set of actions, $T: \mathcal{S}\times\mathcal{A}\times\mathcal{S}\rightarrow[0,1]$ is a transition probability distribution, $R: \mathcal{S}\times\A\times\St \rightarrow \mathbb{R}$ is a reward function, and $\gamma \in (0,1]$ is a discount factor. A solution to an MDP is a policy $\pi: \mathcal{S} \times \mathcal{A} \rightarrow [0,1]$ maximizing expected discounted return $\E \left[\sum_{t=0}^{\infty} \gamma^t r_t\right]$, where $r_t$ is the reward obtained at time step $t$. The value function $V^\pi: \mathcal{S}\rightarrow \mathbb{R}$ for a policy $\pi$ captures the expected return an agent would receive from executing $\pi$ starting in a state $s$. The action-value function $Q^\pi:\mathcal{S}\times\mathcal{A}\rightarrow \mathbb{R}$ of a policy is the expected return from executing an action $a$ in a state $s$ and following policy $\pi$ thereafter.

\textbf{Hierarchical Decision-Making} Solving MDPs with high-dimensional state and action spaces can be difficult, especially in domains where long sequences of actions are required to achieve a goal. In these environments, hierarchical reinforcement learning \cite{barto2003recent} may be more applicable, as temporally-extended actions can reduce the complexity of the space of solution policies. The \textit{options framework} \cite{sutton1999between} formalizes this notion by modeling high-level actions as \textit{options}: closed-loop policies defined by a tuple $(I, \pi, \beta)$, where $I\subseteq\St$ is a set of states in which the option can be executed, $\pi$ is an option policy, and $\beta: \St\rightarrow[0,1]$ describes the probability that the option will terminate upon reaching a given state. If $O$ is the set of options the agent can execute, then the MDP tuple is extended to $(\mathcal{S}, \mathcal{A}\cup O, R, T, \gamma)$ in the hierarchical setting.

\section{RLang: Describing Partial World Knowledge about Tasks}\label{sec:rlang}

If RL is to become widely used in practice, we must reduce the infeasible amount of trial-and-error required to learn to solve a task from scratch. One promising approach is to avoid \emph{tabula rasa} learning by including the sort of background knowledge that humans typically bring to a new task. Such background knowledge is often easy to obtain---in many cases, it is simply obvious to anyone: \emph{try not to fall off cliffs!}---and need not be perfect or complete to be useful.

Unfortunately, however, there is no standardized approach to communicating such background knowledge to an RL agent. In most cases, the same person who implements the learning algorithm also hand-codes the background knowledge, typically in an ad-hoc fashion in the same general-purpose programming language in which the algorithm is implemented. This has two primary drawbacks. First, prior knowledge is often task-specific, and the lack of a means to express it hinders the development of learning algorithms that can exploit varying types and degrees of background knowledge.
Second, it is not accessible to end-users or other consumers of RL agents, who do not write the algorithms themselves and cannot be expected to master the relevant programming languages and mathematical details, but who might nevertheless wish to accelerate learning. 

The alternative is to design a standardized, human-interpretable DSL for expressing prior knowledge about reinforcement learning tasks. Such a DSL should have two important properties not present in existing RL DSLs \cite{maclin1996creating, denil2017programmable, sun2020program}. First, it should be agnostic of the learning algorithm used. Separating the question of how to express prior knowledge from how that knowledge is exploited by a learning algorithm introduces a standardized interface that can be used to inform a wide variety of RL agents, even ones based on algorithms not yet  developed. Second, it should be \emph{complete}: able to express all the information that could possibly be informative about a particular task. We therefore propose \rlang{}, a new DSL designed to fulfill these  criteria.   

RLang can name state features (using \texttt{Features} and \texttt{Propositions}), specify goal states (using \texttt{Goals}), define abstract actions (using \texttt{Options}), describe policies and hierarchical policy structure (using \texttt{Policies}), restrict the action space (using \texttt{ActionRestrictions}), provide partial world models (using \texttt{Effects}), and shape reward (also using Effects). Our Python package parses RLang into an algorithm-agnostic data structure (see Section \ref{sec:knowledge}) that can be integrated into nearly any RL algorithm. 

\begin{table*}[!ht]
\caption{\texttt{RLang} declarations for corresponding MDP elements. The first column shows a component of the MDP, the second shows an \texttt{RLang} expression that can inform it, while the last column contains a description of the expression.}
\small
\centering 
\begin{tabular}{l|p{0.35\linewidth}|p{0.3\linewidth}}
	\toprule
	\bf MDP Component & \bf RLang Declaration & \bf Natural Language Interpretation \\
	\midrule

    \specialcell{State Factor\\
    $\phi: \St\rightarrow\R^n$}
    & 
    \texttt{\textbf{Factor} inventory := \textbf{S}[250:260]}
    & 
    
    Your inventory is a small factor of the state space.
    
    \\
    
    \specialcell{State Feature\\
    $\phi: \St\rightarrow\R^n$}
    & 
    \specialcell{\texttt{\textbf{Feature} inventory_value := }\\\texttt{5 * gold + 2 * iron}}
    & 
    
    The value of your inventory is $5$ for each gold you have plus 2 for each iron.
    
    \\
    
    \specialcell{Proposition\\
    $\sigma: \St\rightarrow\{\True, \False\}$}
    &
    \specialcell{\texttt{\textbf{Proposition} at_workbench :=}\\
    \texttt{{position in workbench_locations}}}
    
    &
    You are at a workbench if your position is one of the workbench locations.
    
    \\

    \specialcell{Objects and Class Definitions\\
    $(\mathcal{C}=\{C_1, ..., C_c\},$\\ $\mathcal{O}=\{o_1, ..., o_n\})$}
    &
    \specialcell{\texttt{\textbf{Class} Block:}\\ \hspace{3mm}\texttt{id: int} \\
    \texttt{\textbf{Object} dirt := Block(1)}}

    &
    Create a Block class and a new dirt Block object.
    \\

    \specialcell{Markov Feature\\
    $f: \St\times\A\times\St\rightarrow\R^n$} 
    &
    \specialcell{\texttt{\textbf{MarkovFeature} inv\_change :=}\\ \texttt{{inventory'} - {inventory}}}
    
    &
    The change in your inventory in a time step

    \\
    \specialcell{Policy\\
    $\pi:\St\times\A\rightarrow[0,1]$} 
    &
    \specialcell{\texttt{\textbf{Policy} build\_bridge:}\\
    \hspace{3mm}\texttt{\textbf{if} at_workbench:}\\ \hspace{3mm}\hspace{3mm}\texttt{\textbf{Execute} {use}}}
    
    &
    If you are at a workbench, craft using it.
    
    \\
    \specialcell{Option\\
    $(\sigma_I, \pi, \sigma_\beta)$}
    &
    \specialcell{\texttt{\textbf{Option} build\_axe:}\\
    \hspace{3mm}\texttt{\textbf{init} (wood >= 1 and iron >= 1)}\\
    \hspace{6mm}\texttt{\textbf{Execute} build\_axe\_policy}\\
    \hspace{3mm}\texttt{\textbf{until} (axe >= 1)}}
    
    &
    When you have at least one wood and one iron, you can build axes until you have at least one.
    \\

    \specialcell{Reward and Transition Function\\
    $(R_e:\St\times\A\times\St\rightarrow\R,$\\
    $T_e: \St\times\A\times\St\rightarrow[0,1])$}
    &
    \specialcell{\texttt{\textbf{Effect} resource_consumption:}\\ \hspace{3mm}\texttt{\textbf{if} wood >= 1 and \textbf{A} == use:} \\
    \hspace{6mm}\texttt{stick' -> stick + wood}\\
    \hspace{6mm}\texttt{wood' -> 0} \\
    \hspace{3mm}\texttt{\textbf{Reward} wood}} 

    &
    Crafting will convert your wood into sticks. You will also be rewarded 1 for every wood you have.
    \\

    \bottomrule
\end{tabular}

\label{tab:elements}
\end{table*}

\subsection{RLang Elements}

An RLang program consists of a set of declarations, where each one grounds to one or more components of an $(\mathcal{S}, \mathcal{A}, O, R, T, \pi)$ tuple. More specifically, every RLang Element grounds to a function with a domain in   $\mathcal{S}\times\mathcal{A}\times\mathcal{S}$ and a co-domain in $\mathcal{S}$, $\mathcal{A}$, $\mathbb{R}^n$ where $n\in \mathbb{N}$, or $\{\True, \False\}$. We describe the main RLang element types in the rest of this section and summarize them in Table \ref{tab:elements}. A full formal semantics and grammar are in Appendix \ref{appendix: semantics}.

\paragraph{State Factors} In Factored MDPs \cite{boutilier2000stochastic}, the state space is a collection of conditionally independent variables: $\mathcal{S} = \mathcal{X}_1 \times .. \times \mathcal{X}_n$. Some algorithms might find it useful to reference these variables individually. For example, consider a 2-D version of Minecraft, as represented in Figure \ref{fig:environments}, where an agent has to collect ingredients to craft new tools and objects. In this environment the state is the concatenation of a position vector, a flattened map representation, and an inventory vector: $s = (\text{pos}, \text{map}, \text{inventory})$. \texttt{Factors} can be used to reference these state variables:

\begin{lstlisting}
Factor position := S[0:2]
Factor map := S[2:250]
Factor inventory := S[250:270]
\end{lstlisting}

\texttt{\textbf{S}} is a reserved keyword referring to the current state. \texttt{\textbf{A}} and \texttt{\textbf{S'}} are also keywords referring to the current action and the next state, respectively. Factors can be sliced and indexed:

\begin{lstlisting}
Factor iron := inventory[0]
Factor wood := inventory[1]
\end{lstlisting}

\paragraph{State Features} RLang can also be used to define more complex functions of state. For instance, if the agent's goal is to build axes, we can define a \texttt{Feature} capturing the number of axes that could be built at the current state:

\begin{lstlisting}
Feature number_of_axes := wood + iron
\end{lstlisting}

\paragraph{Propositions} 
\texttt{Propositions} in RLang, which are functions of the form $\mathcal{S} \rightarrow \{\True, \False\}$, identify states that share relevant characteristics: 

\begin{lstlisting}
Constant workbench_locations := [[1, 0], [1, 3]]
Proposition at_workbench := position in workbench_locations
Proposition have_bridge_material := iron >= 1 and wood >= 1\end{lstlisting}

\paragraph{Goals} \texttt{Goals}  specify goal states via a proposition. For example, \texttt{\textbf{Goal} get_gold := gold >= 1} encodes that the agent must collect at least one gold unit.

\paragraph{Objects and Classes} RLang supports the usage of Object-Oriented MDPs \cite{diuk2008object} via \texttt{Objects} and \texttt{Classes}, with support for sub-classes and various object attribute types (including integers, floats, and booleans). Objects that are in the state space can be accessed, and new objects and classes can be easily instantiated:

\begin{lstlisting}
Class Arm:
    length: int
Object robot_arm := Arm(S.forearm.length + S.end_affector.length)
\end{lstlisting}

\paragraph{Markov Features} Markov functions like the action-value  or transition function take the form $\St\times\A\times \St \rightarrow \mathbb{R}$. We extend the co-domain of this function class to $\mathbb{R}^n$ and introduce \texttt{Markov Features}, which allow users to compute features on an $(s,a,s')$ experience tuple. The following Markov Feature represents a change in inventory elements.

\begin{lstlisting}
Markov Feature inventory_change := inventory' - inventory
\end{lstlisting}

The prime (\texttt{'}) operator references the value of an RLang name when evaluated on the next state.


\paragraph{Policies} Policy functions can also be specified in RLang using conditional expressions:

\begin{lstlisting}
Policy main:
    if iron >= 2:
        if at_workbench:
            Execute Use # This is an action
        else:
            Execute go_to_workbench # This is a policy
    else:
        Execute collect_iron
\end{lstlisting}

The \texttt{Execute} keyword executes an action or calls another policy. The above policy instructs the agent to craft iron tools at a workbench by first collecting iron and then navigating to the workbench. Policies can also be probabilistic:

\begin{lstlisting}
Policy random_move:
    Execute up with P(0.25)
    or Execute down with P(0.25)
    or Execute left with P(0.25)
    or Execute right with P(0.25)
\end{lstlisting}

Users can specify multiple \texttt{policy} functions in an RLang program and designate a primary policy by naming it \texttt{main}.


\paragraph{Options} Abstract actions are specified using \texttt{Options}, which include initiation and termination propositions:

\begin{lstlisting}
Option build_bridge:
    init have_bridge_material and at_workbench
        Execute craft_bridge
    until bridge in inventory
\end{lstlisting}

\paragraph{Action Restrictions} Restrictions to the set of possible actions an agent can take in a given circumstance can be specified using \texttt{ActionRestrictions}:

\begin{lstlisting}
ActionRestriction dont_get_burned:
    if (position + [0, 1]) in lava_locations:
        Restrict up
\end{lstlisting}

\paragraph{Effects} \texttt{Effects} provide an interface for specifying partial information about the transition and reward functions. In a factored MDP, RLang can  specify factored transition functions (i.e., transition functions for individual factors):

\begin{lstlisting}
Effect movement_effect:
    if x_position >= 1 and A == left:
        x_position' -> x_position - 1
    Reward -0.1
\end{lstlisting}

The above Effect captures the predicted consequence of moving left on the \texttt{x_position} factor, stating that the $x$ position of the agent in the next state will be $1$ less than in the current state. This Effect also specifies a $-0.1$ step penalty regardless of the current state or action. In simpler MDPs, predictions can be made about the whole state vector:

\begin{lstlisting}
Effect tic_tac_toe:
    if three_in_a_row:
        S' -> empty_board # Board is reset
\end{lstlisting}

Effects can reference previously defined effects similarly:

\begin{lstlisting}
Effect main:
    -> movement_effect
    -> crafting_effect
\end{lstlisting}

A \texttt{main} Effect designates the primary environment dynamics, and grounds to a partial factored world model $(\overline{\mathcal{T}}, \overline{\mathcal{R}})$. As with policies, Effects can be  probabilistic using \texttt{with}.

Finally, it is important to note that RLang, as we have seen across these examples, does not require the specification of \texttt{Effects} and \texttt{Policies} to be complete. Therefore, a user is not required to provide extensive and complex programs to fully specify the MDP---although this is a possibility with RLang---to accelerate learning. RL agents must learn to solve the task by filling in the missing pieces. 

\subsection{Accessing Parsed RLang Knowledge}\label{sec:knowledge}

Using our Python package, users can parse RLang programs into the following queryable \texttt{knowledge} objects, which can be integrated directly into a learning algorithm:
\begin{itemize}
    \item The \textbf{Dynamics and Task Knowledge} object contains a queryable model of the environment and the task (i.e., transition dynamics $\overline{T}$ and reward function $\overline{R}$) that are derived from the \texttt{\textbf{Effect} main} declaration and the collection of defined goals;
    \item The \textbf{Solution Knowledge} object that contains information about the collection of newly defined options $O$ and the \texttt{main} policy $\overline{\pi}$. 
\end{itemize}
These knowledge objects are implemented as partial functions; an \texttt{unknown} flag is returned when querying for a element of the domain where the RLang  has no knowledge.

\subsection{Selecting a Suitable RLang-informed Agent}

Once an RLang program is compiled into partial MDP components, the parser must select an informed agent that leverages the given information. We propose the following simple mechanism to select among a suite of RLang-informed agents:
Let $\Alpha$ be a set of RL agents capable of taking RLang advice. Let $c_i^\alpha$ be a binary variable that determines whether the agent $\alpha \in \Alpha$ is capable of exploiting a type of advice (e.g., transition dynamics, rewards, etc.). Therefore, for each agent, we have a characterizing vector $C^\alpha = [c_1^\alpha, c_2^\alpha, ..., c_n^\alpha]$. Similarly, the RLang parser generates an analogous binary vector $P$ that represents the RLang information types expressed in the RLang program. Finally, we can select among the agents that exploit as much information as possible by selecting an agent such that $P \approx C^\alpha$.

\subsection{Specifying Complex Groundings with a Vocabulary} 
RLang comes built-in with a set of simple arithmetic, Boolean, and set operations that can be used in RLang object declarations.
However, users may wish to include more complex grounding functions in their RLang programs. For instance, when dealing with problems with high-dimensional observation spaces (e.g., pixel frames), the user may wish to use an object classifier as a means to output a propositional value. Users can therefore import and reference RLang objects defined  using our Python library. By specifying a \textbf{vocabulary file} (in JSON) and a corresponding \textbf{grounding file} (a Python file containing RLang objects), users can construct their own RLang objects and reference them directly in RLang programs. This allows users to provide complex expert groundings or, more generally, learned groundings that hold the necessary semantic information to derive \textit{new} grounded knowledge easily with RLang programs.



\section{Demonstrations}\label{sec:results}

We now provide a few RLang use-cases, focusing on examples that show how different types of prior information can be concisely expressed, and effectively exploited, for varying degrees of environment complexity and different families of RL methods. 
%

\begin{figure}[ht]
    \centering
\begin{subfigure}[b]{0.15\textwidth}
    \includegraphics[width=\textwidth]{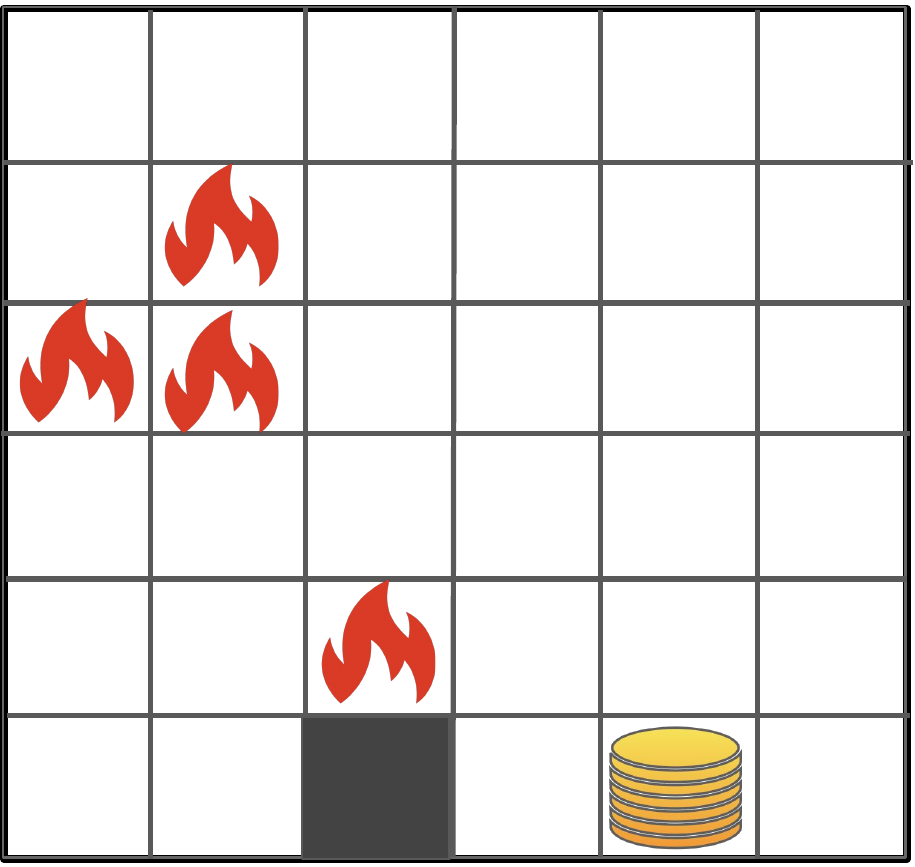}
\end{subfigure} \hspace{20pt}
\begin{subfigure}[b]{0.15\textwidth}
    \includegraphics[width=\textwidth]{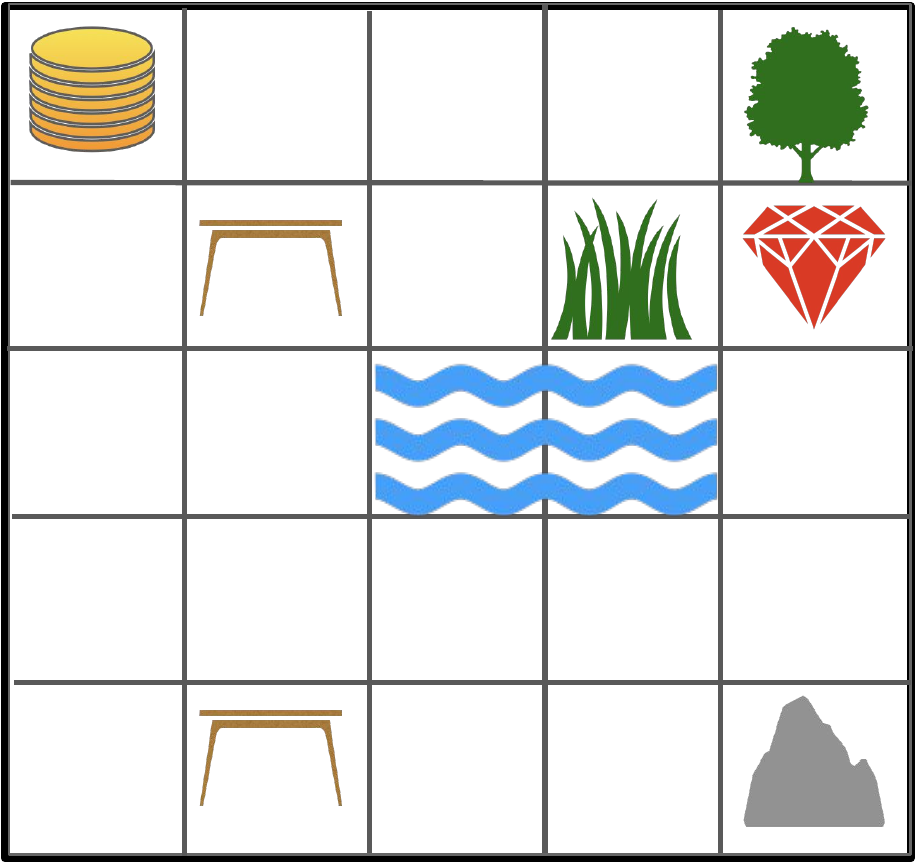}
\end{subfigure}
    \caption{On the left, the Lava-Gap environment. On the right, 2D Minecraft based on \citet{andreas2017modular}.}
    \label{fig:environments}
\end{figure}

\textbf{Hierarchical Policy Structure: 2D Minecraft}
We first consider a 2D version of Minecraft based on \citet{andreas2017modular}, consisting of a gridworld (see Figure \ref{fig:environments}) that contains workbenches where the agent can craft new objects, and raw materials like wood, stone and gold. To build an item, the agent must have the required ingredients and be in the correct workbench. The agent has the action \texttt{use} to interact with elements, and actions to move in the cardinal directions. 

To show how providing the sub-policy structure of the task improves performance, we provide the agent with initiation and termination conditions for a few options (to collect wood, go to the three different workshops and to build the required elements).
The following program concisely defines $3$ options fully and $4$ options with uninformed policies.
\begin{lstlisting}[numbers=left]
    Option go_to_workshop_0:
        init(any):
            Execute go_to_workshop_0_learnable_policy
        until(at_workshop_0)
    Option go_to_workshop_1:
        init(any):
            Execute go_to_workshop_1_learnable_policy
        until(at_workshop_1)
    Option go_to_workshop_2:
        init(any):
            Execute go_to_workshop_2_learnable_policy
        until(at_workshop_2)
    Option get_wood:
        init(there_is_wood):
            Execute get_wood_learnable_policy
        until delta_wood >= 1
    Option build_plank:
        init(wood >= 1 and at_workshop_1):
            Execute use
        until (delta_plank >= 1)
    Option build_stick:
        init (wood >= 1 and at_workshop_1)
            Execute use
        until (delta_stick >= 1)
    Option build_ladder:
        init (stick >= 1 and plank >= 1)
            Execute use
        until (delta_ladder >= 1)\end{lstlisting}
\vspace{0.05cm}
To exploit this information, the agent must learn both the policy over options to maximize reward, and the option policies that achieve each option's termination condition. For both the high-level and low-level agents, we use DDQN  \cite{van2016deep} (details are in Appendix \ref{appendix: minecraft}).

Figure \ref{fig:craftworld-results} show the average return of RLang-informed hierarchical DDQN vs. the uninformed (flat) performance of a DDQN agent. A concise program partially describing a hierarchical solution was sufficient to successfully learn to solve the task, in stark contrast with the uninformed agent. 


\paragraph{Policy Prior: Lunar Lander}
Next, we consider programs that provide prior policy knowledge. Such policy information need not be optimal or complete, but it can still improve learning performance.  
We first consider  Lunar Lander  \citep{openaigym}, which requires learning an optimal control policy to gently land a ship on the moon. The environment has a dense reward encoding both the goal and cost constraints, a continuous state space, and four discrete actions that  either do nothing, fire the main engine, or fire the left or right thruster. 
We provide the agent with an initial policy using the following \rlang{} program:
\vspace{0.05cm}
\begin{lstlisting}[numbers=left]
    Policy land:
        if (left_leg_in_contact == 1.0) or (right_leg_in_contact == 1.0):
            if (velocity_y/2 * -1.0)  > 0.05:
                Execute main_engine
            else:
                Execute do_nothing
        elif remaining_hover > remaining_angle and remaining_hover > -1 * remaining_angle and remaining_hover > 0.05:
            Execute main_engine
        elif remaining_angle < -0.05:
            Execute right_thruster
        elif remaining_angle > 0.05:
            Execute left_thruster
        else:
            Execute do_nothing\end{lstlisting}
\vspace{0.05cm}    
We implemented an \rlang{}-informed agent using PPO \cite{schulman2017proximal}, a policy gradient method, as our base method. 
We probabilistically mixed the \rlang{}-defined advice policy with a learnable policy network using mixing parameter $\beta \in [0,1]$, following \citet{fernandez2006probabilistic}, which is annealed during learning process. In this way, the \rlang{} and the learned policies shared control stochastically. 
Figure \ref{fig:lunarlander-results} shows the average return curves resulting from an uninformed PPO agent \citep{schulman2017proximal} and the \rlang{}-informed version. The informed agent exploits the given policy and learns to improve it further, resulting in a clear performance improvement.

\begin{figure}[ht]
    \centering
     \begin{subfigure}[b]{0.35\textwidth}
         \centering
         \includegraphics[width=\textwidth]{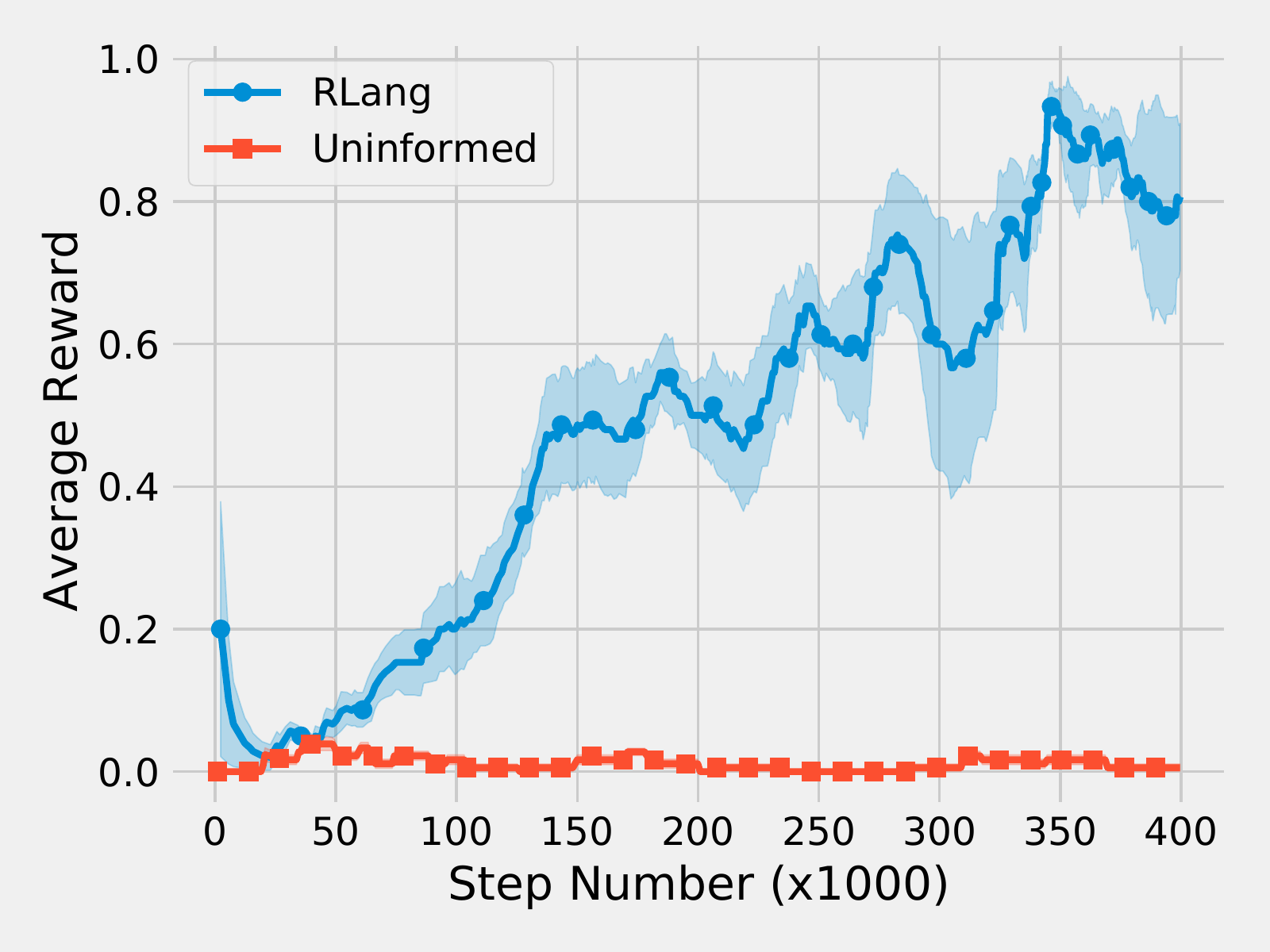}
         \caption{Craftworld + hierarchical information}\label{fig:craftworld-results}
     \end{subfigure}
    \begin{subfigure}[b]{0.35\textwidth}
         \centering
         \includegraphics[width=\textwidth]{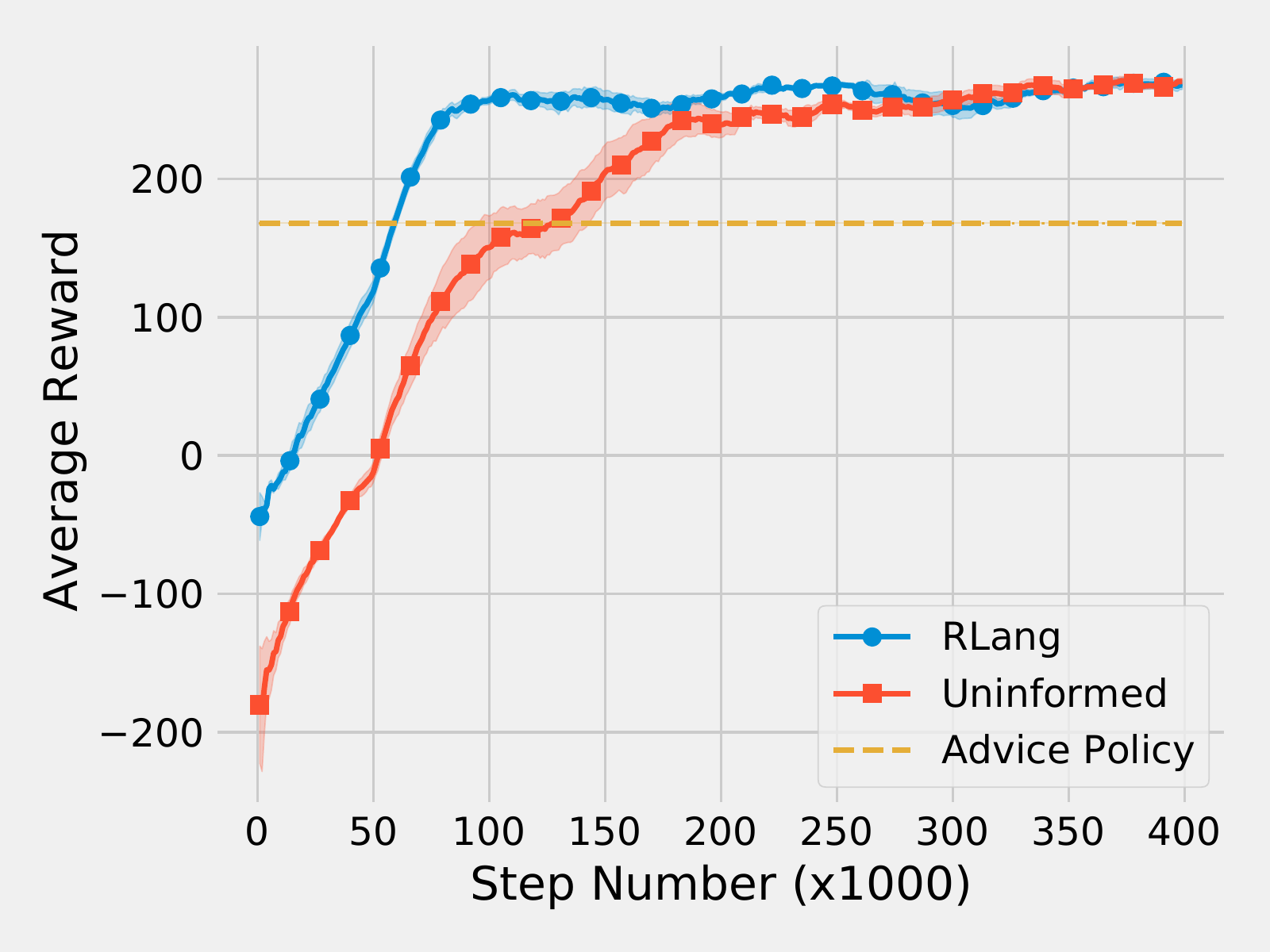}
         \caption{Lunar Lander + policy prior}\label{fig:lunarlander-results}
     \end{subfigure}
    \caption{RLang-informed algorithms given prior \textbf{policy structure}. (a)  an RLang-informed DDQN agent  vs. uninformed DDQN  on 2-D Minecraft. (b) an RLang-informed PPO agent  exploits policy advice vs. learning from scratch.}
    \label{fig:result-subpolicy}
\end{figure}

We also considered two classic control problems: CartPole and Mountain Car. For CartPole, we obtain analogous results using REINFORCE \cite{williams1992simple} as the base method (see Appendix \ref{appendix: classic-control}). In Mountain Car, a hard exploration problem in RL, a very concise \rlang{} policy results in near-optimal performance; the simple program below gets a $-119$ average return over $100$ episodes, where the task is considered solved with a $-110$ average return. 
\vspace{0.25cm}
\begin{lstlisting}[numbers=left]
   Policy gain_momentum:
    if velocity < 0:
        Execute go_left
    else:
        Execute go_right\end{lstlisting}

\paragraph{Dynamics and Rewards: Lava-Gap} 

We now show how to provide transition and reward information using RLang in the Lava-Gap environment (Figure \ref{fig:environments}), a gridworld where an agent must navigate to a goal position. The agent can move in the cardinal directions but each action has a probability of failure of $1/3$. Moreover, moving into walls causes the agent to stay in the same position and falling into a lava pit results in a high negative reward. The agent would typically need to fall into lava pits many times to learn to avoid it. With \rlang{}, however, we can easily inform agents about the dynamics and the high cost of lava pits. 
\begin{figure}[h]
    \centering
     \begin{subfigure}[b]{0.35\textwidth}
         \centering
         \includegraphics[width=\textwidth]{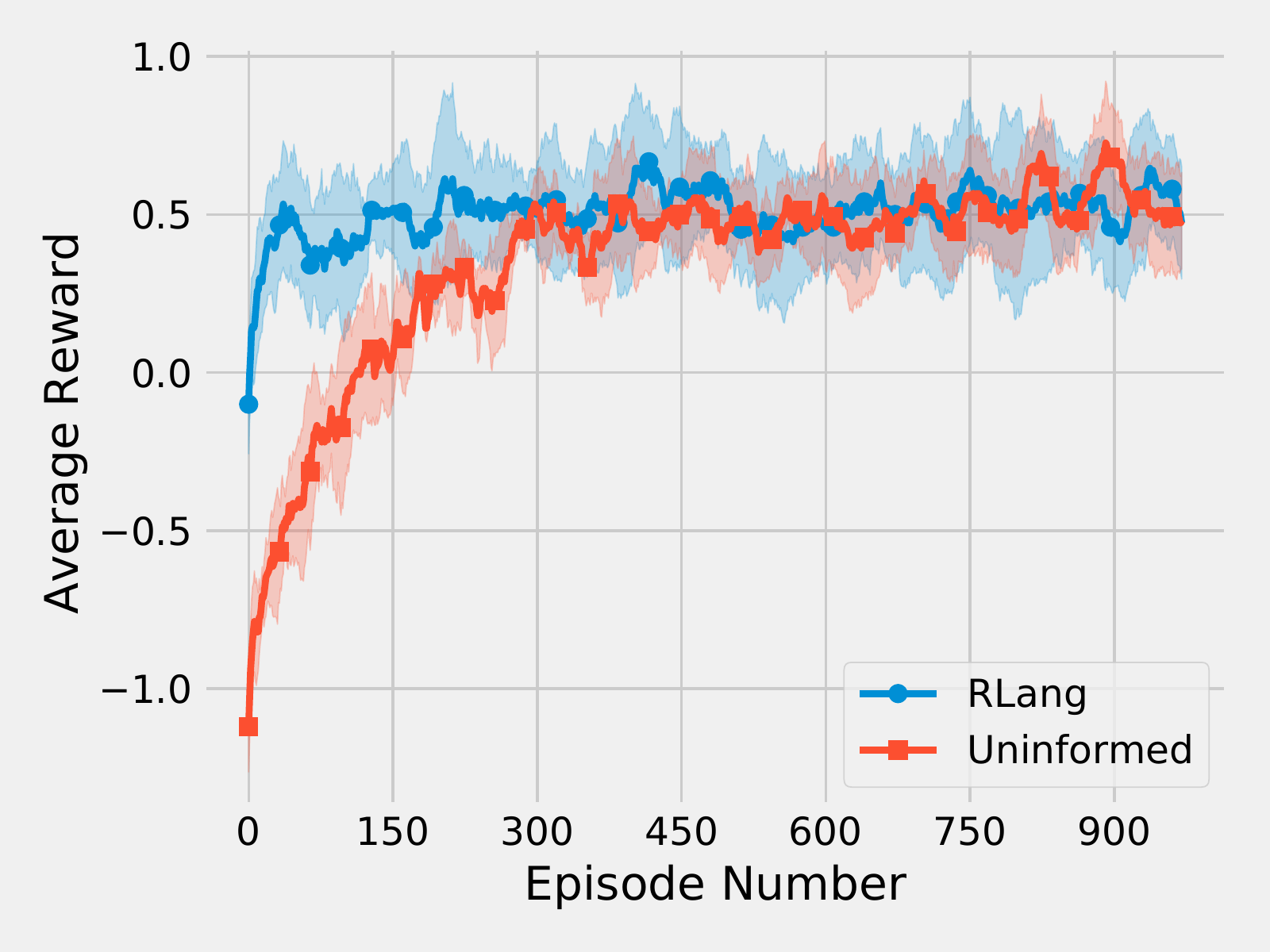}
     \end{subfigure}
    \caption{RLang-informed Q-Learning  agents given partial \textbf{rewards} and \textbf{dynamics} information in Lava-Gap}\label{fig:q-learning-results}
\end{figure}

The program below starts by defining an effect \mbox{\texttt{moving_effect}} that predicts the effect of an action in most cases---i.e., when walls are not in the way. Next, the effect \texttt{dynamics} extends this by describing walls. Finally, the reward function is provided through the effect \texttt{reward}.
\vspace{-0.25cm}
\begin{lstlisting}[numbers=left]
    Effect moving_effect:
        if A == up:
            x' -> x + 1
            y' -> y
        elif A == down:
            x' -> x - 1
            y' -> y
        elif A == left:
            x' -> x
            y' -> y - 1
        elif A == right:
            x' -> x
            y' -> y + 1
    Effect dynamics:  
        if at_wall: 
            S' -> S
        else:
            -> moving_effect
    Effect reward: 
        if in_lava:
            Reward -1
        elif at_goal:
            Reward 1.
        else:
            Reward 0.
    Effect main:
        -> dynamics
        -> reward\end{lstlisting}
Classic tabular Q-Learning is suitable here. We designed a Q-Learning agent that exploits the transition dynamics and reward information. The agent first estimates an initial Q-table using Value Iteration based on the partial transition and reward models---when information is \texttt{unknown} for a transition tuple $(s,a,s')$ the Q-value defaults to $0$. See Appendix \ref{appendix: lava-gap} for more details. 


Average return curves are shown for an RLang-informed Q-Learning agent \cite{watkins1992q} in Figure \ref{fig:q-learning-results}. The curves show that the informed agent clearly leverages the information to gain high return early in the training process.
We obtain analogous results informing an RMax agent \cite{brafman2002rmax}, a model-based method. These results are in Appendix \ref{appendix: lava-gap}.

\textbf{Object-Oriented Dynamics and Rewards: Taxi}

\begin{figure}[h]
    \centering
    \begin{subfigure}[b]{0.35\textwidth}
         \centering
         \includegraphics[width=\textwidth]{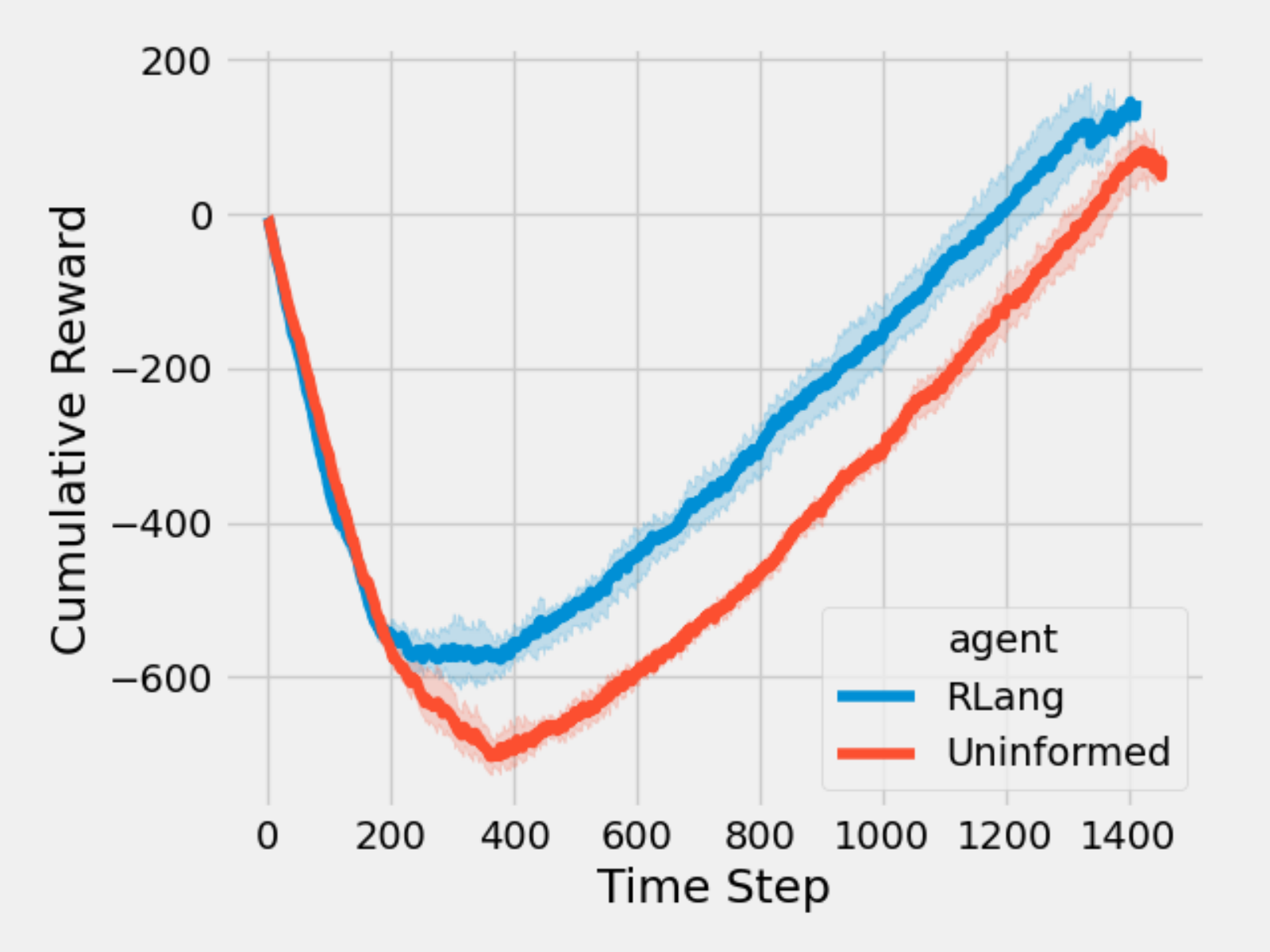}
     \end{subfigure}
    \caption{An RLang-enabled DOORmax agent in object-oriented Taxi informed with dynamics information}\label{fig:taxi-doormax-results}
\end{figure}

We now provide dynamics and reward information for an object-oriented environment using RLang. We use the object-oriented version of the Taxi environment in \citet{diuk2008object}, a gridworld where the agent must pick up a passenger and drop them off at one of four destinations. DOORmax \cite{diuk2008object} is suitable for Object-Oriented MDPs. We implement an RLang-enabled DOORmax algorithm which allows for initialization of transition dynamics and rewards. We show these results in Figure \ref{fig:taxi-doormax-results}.

\begin{lstlisting}[numbers=left]
Effect no_movement_effect:
    if S.taxi.touch_n and A == move_n:
        S' -> S
    if S.taxi.touch_s and A == move_s:
        S' -> S
    if S.taxi.touch_e and A == move_e:
        S' -> S
    if S.taxi.touch_w and A == move_w:
        S' -> S

Effect main:
    if S.taxi.on_passenger and A == pick_up:
        S'.passenger.in_taxi -> True
    if S.passenger.in_taxi and A == drop_off:
        S'.passenger.in_taxi -> False
        if S.taxi.on_destination:
            Reward 20
        else:
            Reward -10
    elif A == pick_up or A == drop_off:
        Reward -10
    else:
        Reward -1
        -> no_movement_effect\end{lstlisting}
\vspace{0.25cm}

OO-MDPs contain an abstract layer of object-centric information about an underlying MDP. RLang provides compact semantics for referencing these abstractions, which object-aware agents can exploit for improved performance. Some predicates given by the Taxi domain describe the taxi's position in the underlying MDP (e.g., \texttt{touch_n(taxi)}), which can be used to specify where the taxi is allowed to move. Using these predicates, the above RLang program succinctly explains that the taxi cannot move through walls, a crucial piece of information that a \textit{tabula rasa} agent might learn for each wall it encounters over many timesteps.



\begin{table*}[ht]
\caption{Comparison of DSLs proposed for RL agents and the types of MDP information that can be expressed}\label{table:related-lang}
\small
\resizebox{\textwidth}{!}{%
    \begin{tabular}{c|ccccccc}
    	\toprule
    	\textbf{RL} & Policy &  Action & Policy  & State & Rewards & Transition\\
    	\textbf{Language} & Hint & Structure & Constraints & Structure & & Dynamics \\
    	\midrule
    	\addlinespace[5pt]
    	ALisp \citep{andre2002state} &  \checkmark  & \checkmark & & & & & \\
    	Advice RL \citep{maclin1996creating} & \checkmark & & \checkmark & & & & \\ 
    	Program-guided Agent \citep{sun2020program} & \checkmark & \checkmark & & & & & \\
    	Programable Agents \citep{denil2017programmable} & & & & \checkmark\\
    	Policy sketches \citep{andreas2017modular} & & \checkmark & & & & &\\
    	GLTL \cite{littman2017environment} & & & & & \checkmark & &\\
    	SPECTRL \cite{jothimurugan2020composable} & & & & & \checkmark & &\\
    	\textbf{RLang} & \checkmark & \checkmark & \checkmark & \checkmark & \checkmark & \checkmark &\\
    	\bottomrule
    \end{tabular}%
    }
\end{table*}

\section{Related Work}

There has been a recent surge of interest in language methods for RL agents \cite{luketina2019survey}. These fall under methods that use natural language to instruct, or to reward, agents as a form of supervision, and methods that use formal languages to represent goals or other components of an MDP.

\textbf{Formal Languages in Reinforcement Learning} In classical planning it is standard to use the Planning Domain Description Language (PDDL; \citealt{ghallab1998pddl}) and its probabilistic extension PPDDL (probabilistic PDDL; \citealp{younes2004ppddl1}) to specify the complete dynamics of a factored-state environment. \rlang{} is inspired by these but it is intended for a fundamentally different task: providing \textit{partial} knowledge to a learning agent, where the knowledge might correspond to any component of the underlying MDP. \citet{maclin1996creating} propose an RL paradigm in which the agent may request advice via a DSL that uses propositional statements to provide policy hints. Similarly, \citet{sun2020program} propose to learn a policy conditioned on a program from a DSL. \citet{andreas2017modular} use a simple grammar to represent policies as a concatenation of primitives (sub-policies) to provide RL agents with knowledge about the hierarchical structure of the tasks.  
Other languages include linear temporal logic (LTL; \citealp{littman2017environment, jothimurugan2020composable}) which has been used to describe goals for instruction-following agents.
These methods ground LTL formulae to reward functions. 
RLang expands on all of these DSLs to include information beyond the policy and the reward function, thus enabling a wider array of information to be provided to the agent.

Table \ref{table:related-lang} summarizes existing DSLs for RL and shows their relative expressive power. Besides RLang, no other DSL is sufficiently powerful to express the wide range of information that could be of use to an RL agent.

\textbf{Natural Language Grounding and Learning Methods}  Several works have attempted to learn mappings for the semantic meaning of natural language to grounded information for agents. 
Some approaches learn to ground the natural language input to a single grounding function type (e.g., reward function) directly from data. 
For instance, RL agents that learn to play Civilization II by grounding linguistic information from manuals as features relevant to estimating the Q-function \citep{branavan2012learning}, and grounding textual specifications of goals and dynamics of the game to learn a language-conditioned policy \citep{zhong2019rtfm}. 
In instruction-following, some approaches learn to map instructions to reward functions \citep{misra2017mapping, bahdanau2018learning, goyal2020pixl2r}. 

However, other approaches translate natural language sentences to an intermediate semantic representation language. In general, these languages are restricted grammars that can be easily mapped to the desired grounding element. 
For example, some methods translate instructions to sequences of primitive actions \citep{misra2015reinforcement} or to LTL formulae \citep{williams2018learning, gopalan2018sequence, patel20}. In future work we plan to use \rlang{} as the semantic representation language, since it is capable of expressing a much wider range of information than existing DSLs. 

\section{Conclusion}

RLang is a concise and unambiguous domain-specific language designed to make it easy for a human to provide background knowledge---about any component of a task---to an RL agent. 
RLang's formal semantics also serve as a unified framework under which to study and compare RL algorithms capable of exploiting background knowledge to improve learning. The examples in this paper show that RLang can be used to provide diverse types of domain knowledge and structure via simple and intuitive RLang programs describing task knowledge that can effectively improve performance over \textit{tabula rasa} methods.

\section*{Acknowledgements}

We would like to thank our colleagues Sam Lobel, Akhil Bagaria, Saket Tiwari and members in the BigAI group for useful conversations, discussions and feedback during the earlier iterations of this project. Also, we thank Michael Littman and Ellie Pavlick whose insights were useful to kick-start this project.
This research was supported in part by NSF grant \#1955361,
NSF CAREER award \#1844960 to Konidaris, and NSF GRFP \#2022339942 to Spiegel, and was conducted using computational resources and services at the Center for Computation and Visualization, Brown University.

\bibliography{ref}
\bibliographystyle{icml2023}
\newpage


\appendix



\newpage
\section{RLang: Grammar and Semantics}\label{appendix: semantics}

In this section, we define the semantics of RLang expressions. 

\subsection{Grammar}
\begin{bnf*}
    \bnfprod{program}{\bnfts{import}  \bnfsp \bnfpn{declarations}}\\
    \bnfprod{declaration}{\bnfpn{constant}}\\
    \bnfmore{\bnfor \bnfpn{action}}\\
    \bnfmore{\bnfor \bnfpn{factor}}\\
    \bnfmore{\bnfor \bnfpn{proposition}}\\
    \bnfmore{\bnfor \bnfpn{goal}}\\
    \bnfmore{\bnfor \bnfpn{feature}}\\
    \bnfmore{\bnfor \bnfpn{markov_feature}}\\
    \bnfmore{\bnfor \bnfpn{object}}\\
    \bnfmore{\bnfor \bnfpn{class_definition}}\\
    \bnfmore{\bnfor \bnfpn{option}}\\
    \bnfmore{\bnfor \bnfpn{policy}}\\
    \bnfmore{\bnfor \bnfpn{effect}}\\
    \bnfmore{\bnfor \bnfpn{action_restriction}}\\
    \bnfprod{constant}{\bnfts{Constant} \bnfsp \bnfpn{identifier} \bnfsp \bnfts{:=}}\\
    \bnfmore{\bnfpn{arithmetic_expression}}\\
    \bnfprod{action}{\bnfts{Action}\bnfsp \bnfpn{identifier} \bnfsp \bnfts{:=}}\\
    \bnfmore{\bnfpn{arithmetic_expression}}\\
    \bnfprod{factor}{\bnfts{Factor}\bnfsp \bnfpn{identifier} \bnfsp \bnfts{:=}}\\
    \bnfmore{\bnfpn{special_variable}}\\
    \bnfprod{proposition}{\bnfts{Proposition}\bnfsp \bnfpn{identifier} \bnfsp \bnfts{:=}}\\
    \bnfmore{\bnfpn{boolean_expression}}\\
    \bnfprod{goal}{\bnfts{Goal}\bnfsp \bnfpn{identifier} \bnfsp \bnfts{:=}}\\
    \bnfmore{\bnfpn{boolean_expression}}\\
    \bnfprod{feature}{\bnfts{Feature}\bnfsp \bnfpn{identifier} \bnfsp \bnfts{:=}}\\
    \bnfmore{\bnfpn{arithmetic_expression}}\\
    \bnfprod{markov_feature}{\bnfts{MarkovFeature}\bnfsp \bnfpn{identifier} \bnfsp \bnfts{:=}}\\
    \bnfmore{\bnfpn{arithmetic_expression}}\\
    \bnfprod{object}{\bnfts{Object}\bnfsp \bnfpn{identifier} \bnfsp \bnfts{:=}}\\
    \bnfmore{\bnfpn{object_instantiation}}\\
    \bnfprod{policy}{\bnfts{Policy}\bnfsp \bnfpn{identifier} \bnfsp \bnfts{:}}\\
    \bnfmore{\bnfpn{policy_statement}}\\
    \bnfprod{policy_statement}{\bnfpn{execute_statement}}\\
    \bnfmore{\bnfor \bnfpn{conditional_policy_statement}}\\
    \bnfmore{\bnfor\bnfpn{probabilistic_policy_statement}}\\
\end{bnf*}
\begin{bnf*}
    \bnfprod{execute_statement}{\bnfts{Execute}\bnfsp}\\
    \bnfmore{\bnfpn{arithmetic_expression}}\\
    \bnfprod{option}{\bnfts{Option}\bnfsp \bnfpn{identifier}\bnfts{:}}\\
    \bnfmore{\bnfpn{option_init}}\\
    \bnfmore{\bnfpn{policy_statement}}\\
    \bnfmore{\bnfpn{option_until}}\\
    \bnfprod{option_init}{\bnfts{init}\bnfsp \bnfpn{boolean_exp}}\\
    \bnfprod{option_until}{\bnfts{until}\bnfsp \bnfpn{boolean_exp}}\\
    \bnfprod{class_definition}{\bnfts{Class}\bnfsp\bnfpn{identifier}\bnfts{:}}\\
    \bnfmore{\bnfpn{attribute_definitions}}\\
    \bnfprod{effect}{\bnfts{Effect}\bnfsp\bnfpn{identifier}\bnfts{:}}\\
    \bnfmore{\bnfpn{effect_statements}}\\
    \bnfprod{effect_statement}{\bnfpn{reward}}\\
    \bnfmore{\bnfor\bnfpn{prediction}}\\
    \bnfmore{\bnfor\bnfpn{effect_reference}}\\
    \bnfmore{\bnfor\bnfpn{conditional_effect}}\\
    \bnfmore{\bnfor\bnfpn{probabilistic_effect}}\\
    \bnfprod{reward}{\bnfts{Reward}\bnfsp}\\
    \bnfmore{\bnfpn{arithmetic_expression}}\\
    \bnfprod{prediction}{\bnfpn{identifier}\bnfts{'}\bnfsp\bnfts{->}\bnfsp }\\
    \bnfmore{\bnfpn{arithmetic_expression}}\\
\bnfprod{effect_reference}{\bnfts{->}\bnfsp \bnfpn{identifier}}\\
    \bnfprod{action_restriction}{\bnfts{ActionRestriction}\bnfsp}\\
    \bnfmore{\bnfpn{identifier}\bnfts{:}}\\
    \bnfmore{\bnfpn{restrict_statements}}\\
\end{bnf*}

\subsection{Semantics: Basic Syntactic Elements}

RLang allows to express information that grounds to functions defined over the State-Action space of an MDP. Moreover, these functions have to be Markov, in the MDP sense, allowing to define functions with domain $X$ that can be $\St\times\A\times\St$ and, its simplifications, $\St$, $\A$, $\St\times\A$, $\St\times\St$. The range of these functions can include real vectors $\R^d$ (with $d \in \mathbb{N}$), Booleans $\{\True, \False\}$ and sets. The following are the basic expressions that are used to build the MDP specific elements of RLang.

\begin{itemize}

     \item \textbf{Real Expressions} $\bnfpn{arithmetic_expression}$ are functions of the form $f: X \rightarrow \R^d$ for some dimension $d$. Syntactically, RLang allows element-wise arithmetic operations ($+, -, *, /$), numeric literals, and references to previously defined real functions to define new functions. These functions are Markov and defined over the State-Action Space of the MDP, i.e. $X \in \left\{ \St, \A, \St\times\A, \St\times\St, \St\times\A\times\St \right\}$.
     
    \item \textbf{Constant expressions} $\bnfpn{constant}$ allows to bind a name ($\bnfpn{identifier}$) to literal value or a list of literal values.
    
     
    \item \textbf{Boolean Expressions}  $\bnfpn{boolean_expression}$ Analogous to Real Expressions, these are functions of the form $f: X\rightarrow\left\{\True, \False\right\}$ with domain  $X \in \left\{ \St, \A, \St\times\A, \St\times\St, \St\times\A\times\St \right\}$.

    In order to define new Boolean expressions, RLang allow for logical operators \texttt{(and, or, not)} and order relations of the real numbers (<, <= , >, >=, =, !=). 
     
    \item \textbf{Relations and Partial Functions} A relation of domains $X$ and $Y$ are a subset $R\subseteq X\times Y$. Partial functions are specifications of functions of the form $f: X \rightarrow Y$. A partial function is, then, a relation such that
    \begin{align*}
         PF := &\{ (x,y) \in X\times Y \text{ and }\\
          &\forall (x, y), (x', y') \text{  } x = x' \implies y = y' \}.
    \end{align*}
    
    Therefore, it is partial because not every element of the domain is defined by the function. We consider that undefined domain elements map to the special element \texttt{unknown}.

     
     
     
\end{itemize}

\subsection{Semantics: RLang Expressions}

    \textbf{Core RLang Type Definitions} are necessary in order to derive new information from the base vocabulary. 
    
    \begin{itemize}
        \item \textbf{State Space Definitions} allows to define important features and set of states for the agent.
        
        \begin{enumerate}
            \item \textbf{State Features} $\bnfpn{feature}$ These ground to functions of the form $\phi: \St \rightarrow \R^d$. Hence, given the base vocabulary and using real expressions, we can derive new features;
            \item \textbf{State Factors} $\bnfpn{factor}$ In the particular case the state space is $\St \subseteq \R^n$ ($n \in \mathbb{N}$) and factored, we have specific state features that correspond to the State Factors whose definition is given by a list of integers that correspond to the positions in the state vectors that are part of the factor. A factorization of the state correspond to a set of disjoint factors whose union is the full state vector;
            \item \textbf{Propositions} $\bnfpn{proposition}$ ground to Boolean functions with domain in $\St$ that represents a set of states $S_\sigma := \{ s\in\St \text{ and } s\models \sigma \}$;
        \end{enumerate}
        
        \item \textbf{Action Definition} $\bnfpn{action}$ names a particular action. Consider that the action space $\A \subseteq \R^d$ with $d \in \mathbb{N}$. Then, an action definition grounds to a point in $\A$. 
        
        \item \textbf{Option Definition} $\bnfpn{option}$ allows to define an temporally-abstracted action based on the options framework \cite{sutton1999between}. Therefore, the statement directly maps to the triple $(I, \pi_o, \beta)$, where $I$ is the initiation proposition defined by the syntactical element $\bnfpn{option_init}$, $\beta$ is the termination proposition defined by $\bnfpn{option_until}$ and $\pi_o$ is the policy function defined by a policy statement $\bnfpn{policy_statement}$

        \item \textbf{Class and Object Definition} $\bnfpn{class}$ and $\bnfpn{object}$ allow for the definition of new classes and object instances. Classes can have any number of attributes of any of the types \texttt{str}, \texttt{int}, \texttt{float}, \texttt{bool}, \texttt{object} and object instances can have attributes that are functions of the $(s,a,s')$ tuple.
        
        \item \textbf{Markov Feature Definition} $\bnfpn{markov_feature}$ allow for the definition of real-valued features of a transition tuple $(s,a,s')$ of the MDP. These expressions ground to real functions of the form $f: \St\times\A\times\St \rightarrow \R^n$ with $n \in \mathbb{N}$.
    \end{itemize}
    
    \textbf{Core MDP Expressions} are related to specifications of the main functions of the MDP:

    \textbf{Transition Dynamics and Rewards} represented syntactically by $\bnfpn{effect_statement}$. Such statements ground to tuples $(R, T)$ of reward functions and next-state probabilities. More precisely, grounding functions are $R: \St\times\A\times\St\rightarrow\R$ and $T: \St\times\A\times\St\rightarrow[0,1]$. The following are the semantics of possible effect expressions.
        
        \begin{itemize}
            \item \textbf{Reward} $\bnfpn{reward}$ statement allows to specify a reward value using real expressions. A reward statement grounds to a function $R: \St\times\A\times\St \rightarrow \R$ defined by scalar arithmetic expressions.
            
            \item \textbf{Next State Prediction} $\bnfpn{prediction}$ ground to functions $T:\St\times\A\times\St\rightarrow[0,1]$ that gives a probability of transitioning to the next state $s'$ after executing action $a$ at state $s$. The following are possible groundings:
            
            \begin{enumerate}
                \item \textbf{Null Effect}: \texttt{\textbf{S'} -> \textbf{S}}. This grounds to 
                    \begin{equation*}
                        T(s', a, s) = \begin{cases}
                            1 & \text{if } s' = s\\
                            0 & \text{otherwise}
                        \end{cases};
                    \end{equation*}
                \item \textbf{Singleton Prediction}: \texttt{\textbf{S'} -> $\bnfpn{constant}$}. Let the constant ground to a valid state vector $\hat{s}\in\St$. Thus, the expression grounds to
                \begin{equation*}
                        T(s', a, s) = \begin{cases}
                            1 & \text{if } s' = \hat{s}\\
                            0 & \text{otherwise}
                        \end{cases};
                    \end{equation*}
                \item \textbf{Real Prediction}: \texttt{\textbf{S'} -> $\bnfpn{arithmetic_expression}$}. Let the $\bnfpn{arithmetic_expression}$ ground to a real function of the form $e: \St\times\A \rightarrow \St$. Then, the prediction expression grounds to
                \begin{equation*}
                        T(s', a, s) = \begin{cases}
                            1 & \text{if } s' = e(s,a)\\
                            0 & \text{otherwise}
                        \end{cases};
                    \end{equation*}
                \item \textbf{Factor Prediction} \texttt{factor_name -> $\bnfpn{arithmetic_expression}$}. Let \texttt{factor_name} ground to the factor $\phi: \St \rightarrow \R^d$ where $d\in\mathbb{N}$ is the dimension of the factor. Let the $\bnfpn{arithmetic_expression}$ ground to $e_\phi: \St\times\A\rightarrow\R^d$. Then, a factored prediction grounds to the function
                
                \begin{equation*}
                        T_\phi(\phi(s'), a, s) = \begin{cases}
                            1 & \text{if } \phi(s') = e(s,a)\\
                            0 & \text{otherwise}
                        \end{cases};
                \end{equation*}
                
                A collection of factor predictions for a set of disjoint factors $\{\phi_i\}_i$ that partition the state vector can ground to a full transition function
                
                \begin{equation*}
                    T(s',a,s) = \prod_i T_{\phi_i}(\phi_i(s'), a,s)
                \end{equation*}

                \item \textbf{Probabilistic Effect Statements} $\bnfpn{probabilistic_effect_statement}$ allows to explicitly indicate probabilities for a collection of predictions. Consider that the probabilistic statement is a collection of tuples $\{(T_i, p_i)\}_i$ where $T_i$ is the grounding function of the prediction and $p_i$ a probability, subject to the correctness of the probabilities $\sum_i p_i \leq 1$ and for all $p_i \geq 0$.
                Thus, it grounds to
                \begin{equation*}
                    T(s',a,s) = \sum_i p_iT_i(s',a,s).
                \end{equation*}
                
                If $\sum_i p_i < 1$, then the remaining probability is construed to be assigned to \texttt{unknown}.
                
                In the case of Factor predictions for a given factor, the grounding function $T_\phi$ is defined analogously.
                
                \item \textbf{Conditional Effect Statements} $\bnfpn{conditional_effect_statement}$ The conditional context allows to define subsets of the domain $D \subseteq \St\times\A\times\St$ through a $\bnfpn{boolean_expression}$ that defines when a particular \texttt{Execute} statement is valid. Hence, a $\bnfpn{conditional_policy_statement}$ grounds to  partial functions $R = \{(D_i, R_i) \}_i$ where each tuple is the result of a branch from the parsing of \texttt{if-elif-else} blocks. Analogously, for the grounding of dynamics information of the statement $T = \{(D_i, T_i) \}_i$, where the $D_i$ are disjointed subsets of the domain defined by the Boolean expressions, the order of the conditional branches and the $T_i$ and $R_i$ are defined by reward and prediction statements defined above.
                
                \item \textbf{Effect References} $\bnfpn{effect_reference}$ allows to refer to previously defined effect statements to compose new ones. Each referred effect is tuple of $(R_i, T_i)$ of groundings for rewards and transition dynamics. 
                
                A collection of effect references ground to:
                
                \textbf{Rewards} $R(s,a,s') = \sum_{i\in I(s,a,s')} R_i(s,a,s')$ where $I(s,a,s')$ is the set of referred effects that have information for rewards in the tuple $(s,a,s')$. Hence, rewards are composed additively. 
                
                \textbf{Transition Dynamics} Let $S_i'(s,a) = \{s' \in \St : T_i(s',a,s) > 0\}$ be the set of next states from state-action pair $(s,a)$ about which $T_i$ provide knowledge. Thus, a set of effect references are well-defined if $\bigcap_i S_i'(s,a) = \emptyset$ and $\sum_{s'\in \bigcup_i S_i'(s,a)} T(s',a,s) \leq 1$ for all $(s,a)$. Hence, the grounding function is
                \begin{equation*}
                    T(s',a,s) = \sum_i{T_i(s',a,s)}.
                \end{equation*}
                
            \end{enumerate}
            
        \end{itemize}

        \textbf{Policies} $\bnfpn{policy}$ ground to policy functions $\pi : \St\times\A \rightarrow [0,1]$. The simplest expression to specify a policy is an \textbf{execute statement} $\bnfpn{execute_statement}$. The name after the \texttt{Execute} keyword represents either an $\bnfpn{action}$ $a' \in \A$ and, hence, the statement grounds to 
        
        \begin{equation*}
            \pi(s,a) = \begin{cases}
                1 & \text{if } a = a'\\
                0 & \text{otherwise}
            \end{cases},
        \end{equation*}
        
        or it can refer to a previously defined $\bnfpn{policy}$ that grounds to $\hat{\pi}$ and, then, the statement grounds to $\pi = \hat{\pi}$. Therefore, the $\bnfpn{execute_statement}$ functions analogously to a return statement in function definitions in imperative programming languages: when an action name is found, it maps the querying state $s$ to the first action referenced by an execute statement.
        
        \textbf{Probabilistic policy expressions} $\bnfpn{probabilistic_policy_statement}$: Probability statements allow to extend the execute statement with explicit probability values. Therefore, a probabilistic policy statement grounds to a collection of execute statements-probability pairs $\{(\pi_i, p_i)\}_i$. In this way, probabilistic policy statements ground to
        
        \begin{equation*}
            \pi(s,a) = \sum_i p_i \pi_i(s,a)
        \end{equation*}
        
        If $\sum_i p_i < 1$, then the remaining probability is construed to be assigned to \texttt{unknown}.
        
        \textbf{Conditional Policy Expressions} $\bnfpn{conditional_policy_statement}$: The conditional context allows to define subsets of the domain $S' \subseteq \St$ through a $\bnfpn{proposition}$ that defines when a particular execute statement is valid. Hence, a $\bnfpn{conditional_policy_statement}$ grounds to a partial function $\pi' : \{(S_i, \pi_i) \}_i$ where each pair $(S'_i, \pi_i)$ the result of a branch from the parsing of \texttt{if-elif-else} blocks. The $S'_i$ are disjointed and they are the result of the $\bnfpn{proposition}$, the order of the statements and the returning semantics of the execute statements. The $\pi_i$ are defined by execute statements or probabilistic policy statements.
        
        \textbf{Action Restrictions} $\bnfpn{action_restriction}$ are defined analogously to conditional policy statements. They reduce the possible set of actions to consider in a given situation. They ground to functions of the form $A: \St \rightarrow \A$ that defines then subset of prohibited actions to take in state $s$---i.e., $A(s) \subset \A$.
        
        \textbf{Goals} $\bnfpn{goal}$ ground to set of states that are considered goal states for the MDP, i.e. terminating and highly rewarding. RLang represents goals through propositions.

\section{Experimental Details and Additional Results}
In this section, we extend the discussion on the implementation details of RLang demonstrations in Section \ref{sec:results}. We provide details about the implementation of the RLang-informed variations of the RL algorithms, descriptions of the environments and the hyperparameters used.

For each of the experiments below we report average return curves over $5$ different random seeds and report $95\%$ confidence intervals. Moreover, we use a running average with window size of $50$.

\begin{figure*}[!ht]
    \centering
    \begin{subfigure}[b]{0.4\textwidth}
         \centering
         \includegraphics[width=\textwidth]{images/lavaworld/QLearning-average-reward.pdf}
         \caption{RLang-informed Q-Learning}
     \end{subfigure}
    \begin{subfigure}[b]{0.4\textwidth}
         \centering
         \includegraphics[width=\textwidth]{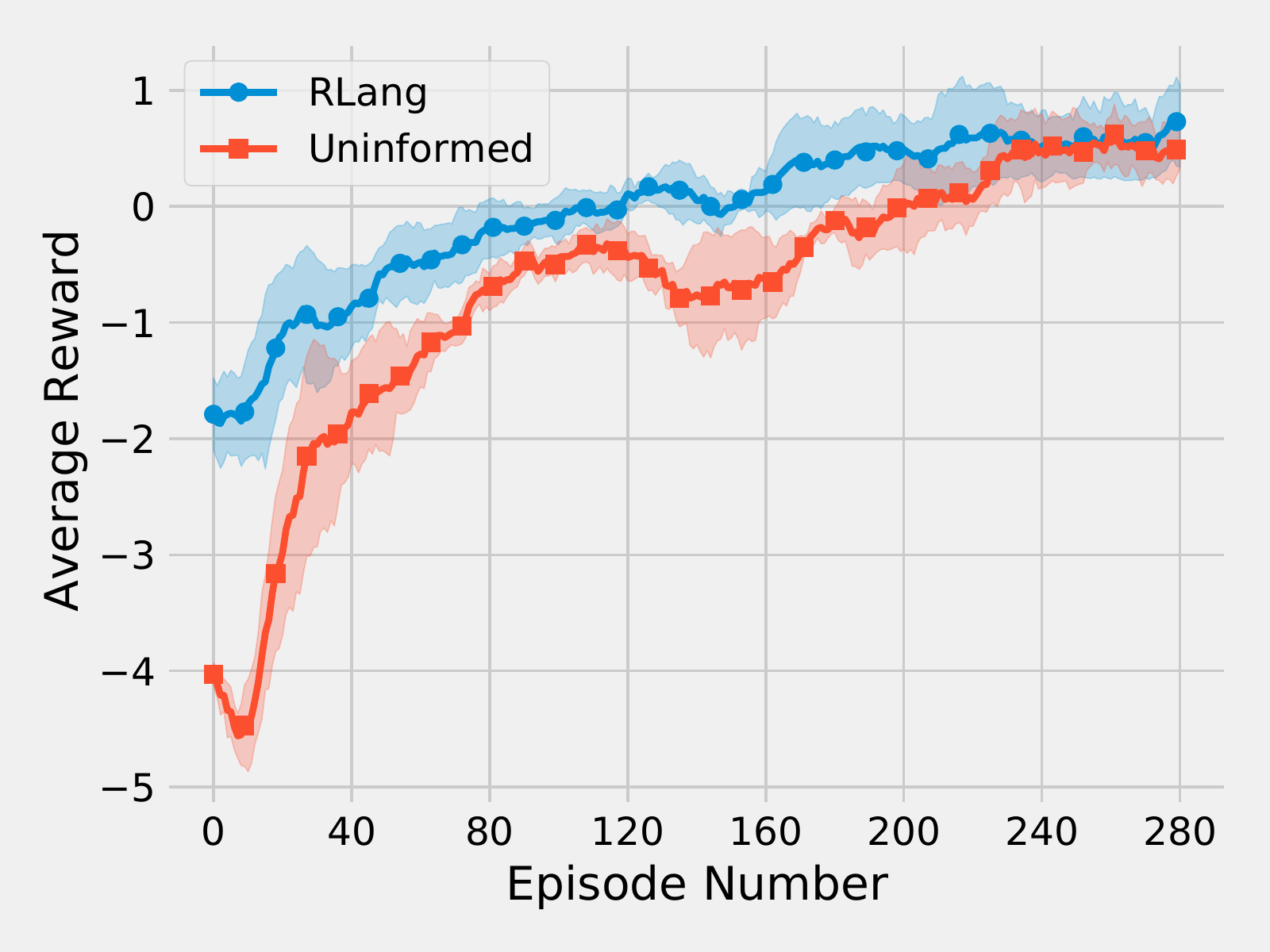}
         \caption{RLang-informed RMax}\label{fig: rmax-results}
     \end{subfigure}
    \caption{Average return curves for Lava-Gap environment. We provide the agent with an RLang program that contains information about the reward function and the transition dynamics.} 
\end{figure*}

\subsection{Lava-Gap} \label{appendix: lava-gap}

Lava-Gap is a $6\times 6$ grid-world with coordinates $x, y \in $ $\{1, 6\}$. There is a wall in position $(3,1)$ and $4$ lava pits in locations $(3,2), (1,4), (2,4), (2,5)$. The goal position is $(5,1)$. The agent has $4$ discrete actions that allows it to move in one of the cardinal directions by $1$ step. Each action has a probability of failure of $1/3$ that would move the agent to a random neighboring position (in the cardinal directions). The state $s_t$ at time $t$ is represented by the $x$ and $y$ coordinates of the position at time $t$. The agent receives a reward of $-1$ for falling in a lava pit and the episode terminates. Similarly, the agent receives a reward of $1$ for reaching the goal and the episode terminates. In any other case, the reward is $0$. At the start of every episode, the agent begins executing at position $(1,1)$. We use a discount factor of $\gamma=0.95$.
We use \texttt{simple\_rl}'s implementation \cite{abel2019simplerl} of this gridworld and of the RMax and Q-Learning algorithms.

\paragraph{RLang-informed Q-Learning}
In the RLang-informed example for Lava-gap, we leverage the transition and reward information from the RLang program to initialize the Q-table for Q-Learning and R-Max. We compute the Q-table by executing value iteration considering in the state pairs where transition information is available. In Algorithm \ref{alg:qtable-init}, we show how the Q-table of a Q-Learning agent is initialized using the information provided in a RLang program given as input as the RLangKnowledge objects. The UpdateValue function computes the new value using the standard TD-error.

\begin{algorithm}
\caption{Q-Table Initialization} \label{alg:qtable-init}
\begin{algorithmic}[0]
\FUNCTION {InitQTable(RLangKnowledge, QAgent)}
\FOR{(s,a) $\in \St\times\A$}
\IF{RLangKnowledge.Reward(s,a) is \texttt{known}}
    \STATE QAgent.Q(s,a) $\gets$ RLangKnowledge.Reward(s,a)
\ENDIF
\ENDFOR
\FOR {$N$ iterations}
\FOR{(s,a,s') $\in \St\times\A\times\St$}
    \IF{RLangKnowledge.Transition(s,a) is \texttt{known}}
    \STATE $T(s,a,s')\gets \text{RLangKnowledge.Transition}(s,a)$
    \STATE $R(s,a,s') \gets \text{RLangKnowledge.Reward(s,a)}$
    \STATE QAgent.Q(s,a) $\gets \text{UpdateValue}(s,a,s', T(s,a,s'), \text{QAgent})$
    \ENDIF
\ENDFOR
\ENDFOR
\ENDFUNCTION
\end{algorithmic}
\end{algorithm}

\paragraph{RLang-informed RMax}
RMax \cite{brafman2002rmax} is a model-based RL algorithm. Hence, we directly use the RLang-provided information to initialize the model of the world (i.e., Transition and Reward tables) in addition to initializing the Q-table. We set the count in the respective tables to be a hyper-parameter $K$, less that the threshold used by RMax to considered a transition known. 

\paragraph{Hyperparameters} For uninformed Q-Learning we have the exploration $\epsilon = 0.1$ and the step size $\alpha = 0.05$ and for the RLang-informed Q-Learning we use $\epsilon=0.01$ and the same $\alpha$. 
For RMax and RLang-informed RMax, we use a threshold of $30$ samples to consider the transition learned and set $K=1$.

\paragraph {RMax Results} In Figure \ref{fig: rmax-results}, we show the average return curves for an RMax agent informed with the program below, in which we see consitent results with the results obtained for an RLang-informed Q-Learning agent.

            
    
        
\subsection{Taxi (Flat)} 

We use \texttt{simple\_rl}'s implementation of the Taxi environment \cite{dietterich1998maxq} with $2$ passengers in a $5\times 5$ grid. The state vector has the position of the agent and a binary variable that is $0$ when the taxi does not carry a passenger and $1$ otherwise. Moreover, it has the current position of the passengers, the destination of the passenger and a binary variable that indicates if the passenger is in the taxi. The agent has $4$ movement actions and a special action for picking up a passenger that is at the same position than the agent and another for dropping off the passenger currently in the taxi. The reward function is $1$ when all passenger are in destination and $0$ otherwise. The discount factor $\gamma=0.95$.

\paragraph{RLang-informed hierarchical Q-Learning} In this experiment, we use a hierarchical RL agent based on the options framework. In this particular case, we use Q-Learning to learn both the policy over options and the intra-option policies. We consider that an RLang-defined option is learnable if the policy function is not provided, i.e. only initiation and termination conditions are specified. We use such termination condition as a goal represented by a pseudo-reward function that is $1$ when the termination condition is achieved and $0$ otherwise that the inner agent uses to learn the intra-option policy. We initialize the intra-option learning agents of those learnable options defined in the input RLang program with the procedure in Algorithm \ref{alg:hrl-init}.

\begin{algorithm}[h]
\caption{Hierarchical Agent Initialization} \label{alg:hrl-init}
\begin{algorithmic}[1]
\FUNCTION {InitializeOptions(RLangKnowledge)}
\FOR{$o\in \text{RLangKnowledge.Options}$}
    \IF{$o$ is \texttt{learnable}}
        \STATE $o.\text{agent} \gets \texttt{InitializeAgent()}$
    \ENDIF
\ENDFOR
\ENDFUNCTION
\end{algorithmic}
\end{algorithm}

\paragraph{Hyperparameters} For our Q-Learning baseline, we use an exploration $\epsilon=0.1$ and a step size of $\alpha = 0.1$. For our hierarchical Q-Learning agents, we have a Q-Learning agent for each subpolicy to be learnt with $\epsilon=0.1$ and $\alpha=0.1$ and a Q-Learning agent to learn the policy over options with $\epsilon=0.01$ and $\alpha=0.5$. We implemented our hierarchical Q-Learning agent as \citeauthor{sutton1999between}.

\paragraph{Results} In Figure \ref{fig: taxi-results}, we show the average return curves for Taxi. The RLang program, shown below, defines the options given to the agent---a simple variation of this program is provided to the agent that needs to learn the intra-option policies.
The plot shows the most-informed agent, i.e. intra-option policies are provided and it only needs to learn the policy over options, is represented in blue, the RLang-informed agent that only knows the initiation and termination conditions of the options (in red) and an uninformed Q-Learning agent. We observe that both of the informed agents are able to exploit the knowledge to gain a steeper learning curve with respect to the uninformed agent.

\begin{lstlisting}[numbers=left]
    Option pick_up_passenger0:
        init(not(passenger_in_taxi) and not(passenger_0_in_dest))
            Execute pick_up_passenger0
        until passenger_0_intaxi

    Option dropoff_passenger0:
        init(passenger_0_intaxi)
            Execute dropoff_passenger0
        until passenger_0_in_dest and not(passenger_0_intaxi)
    
    Option pick_up_passenger1:
        init(not(passenger_in_taxi) and not(passenger_1_in_dest))
            Execute pick_up_passenger1
        until passenger_1_intaxi
    
    Option dropoff_passenger1:
        init(passenger_1_intaxi)
            Execute dropoff_passenger1
        until passenger_1_in_dest and not(passenger_1_intaxi)\end{lstlisting}

\begin{figure*}[ht]
    \centering
    \begin{subfigure}[b]{0.3\textwidth}
         \centering
         \includegraphics[width=\textwidth]{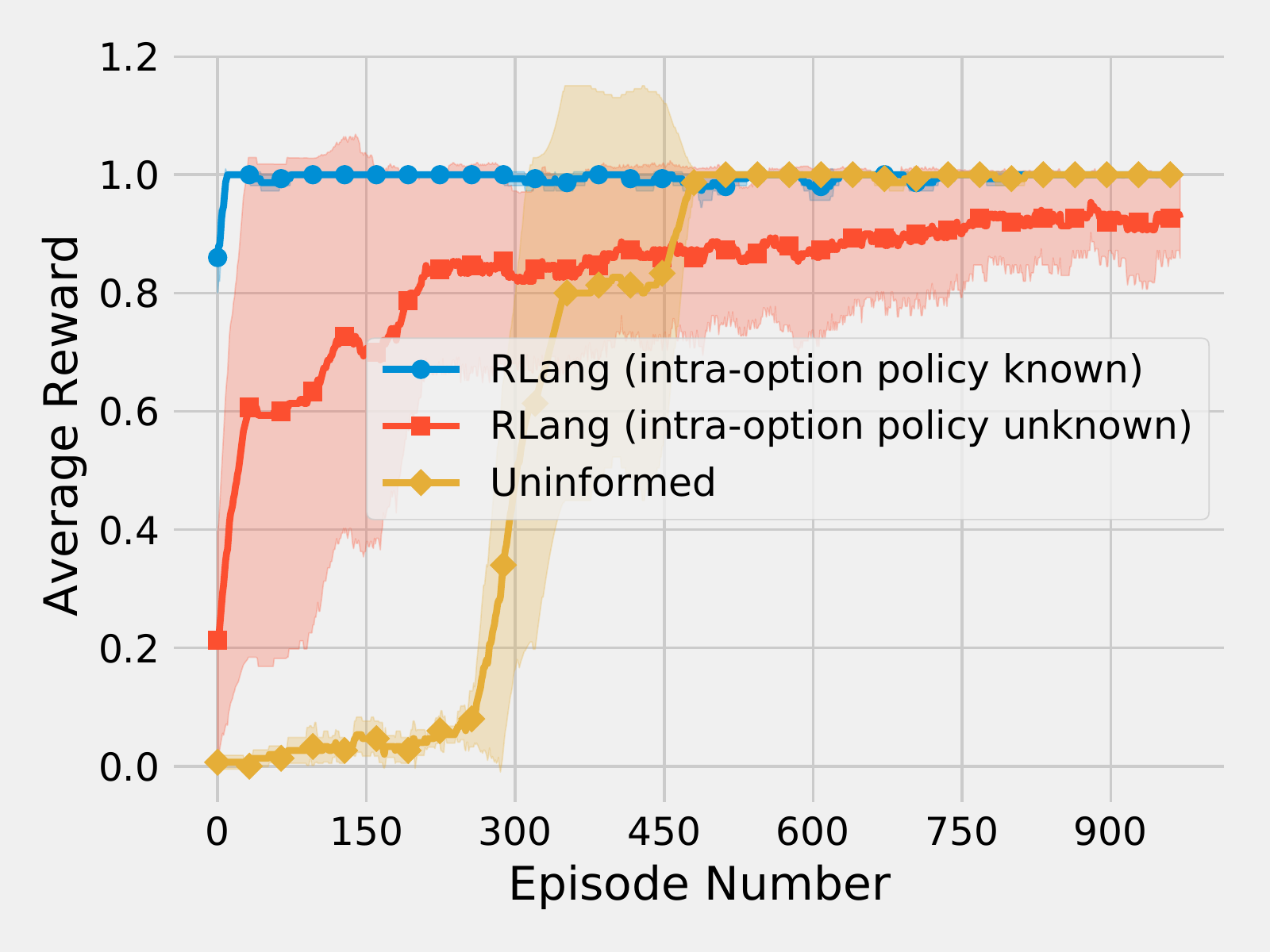}
         \caption{Taxi (Flat)}\label{fig: taxi-results}
     \end{subfigure}
     \begin{subfigure}[b]{0.3\textwidth}
         \centering
         \includegraphics[width=\textwidth]{images/taxi-oo/cumulative-reward.pdf}
         \caption{Taxi (Object-Oriented)}\label{fig: taxi-oo-results-appendix}
     \end{subfigure}
    \begin{subfigure}[b]{0.3\textwidth}
         \centering
         \includegraphics[width=\textwidth]{images/craftworld/craftworld-mean-reward.pdf}
         \caption{2D Minecraft}
     \end{subfigure}
    \caption{Cumulative reward curves for object-oriented Taxi (middle) when the reward function and partial transition dynamics is provided using RLang. Average return curves for flat Taxi (left) and Craftworld (right) when information about the hierarchical structure of the problem is provided using RLang. We provide the agents with a program that specifies the initiation and termination conditions of the options required to solve the problem and the RLang-informed agent learns the intra-option policies and policy over options. In the particular case of flat Taxi, we also include the learning curve when we also provide the intra-option policies.}
\end{figure*}

\subsection{Taxi (Object Oriented)}
For object-oriented Taxi we used \texttt{efficient_rl}'s implementation of the environment presented in \citet{diuk2008object} with 1 passenger in a $5 \times 5$ grid. The agent has access to the same information as the flat Taxi environment with the addition of a set of seven predicates whose truth values are available at each state:

\begin{lstlisting}
touch_north(taxi, wall)
touch_south(taxi, wall)
touch_east(taxi, wall)
touch_west(taxi, wall)
on(taxi, passenger)
on(taxi, destination)
in_taxi(passenger)
\end{lstlisting}

The agent's action space is the same as the flat Taxi environment: 4 movement actions and a special action for picking up a passenger. The reward function is different than flat Taxi, however, and matches one given in \citet{diuk2008object}: a reward of $20$ for dropping off a passenger at a depo, a reward of $-10$ for dropping off a passenger anywhere else or attempting a pickup when the taxi is not on a passenger, and a reward of $-1$ otherwise. The discount factor is likewise $\gamma = 0.95$.

\paragraph{RLang-informed DOORmax} In this experiment we use the DOORmax algorithm implemented by \texttt{efficient_rl} that was originally presented in \citet{diuk2008object} as a baseline and create an RLang-enabled DOORmax agent. Similar to the RLang-enabled Q-Leaning agent, the RLang-enabled DOORmax agent starts by initializing its internal representation of the transition and reward function with the partial dynamics given by an RLang program.

\paragraph{Hyperparameters} Both DOORmax and RLang-DOORmax use the same hyperparameters. We set $r_{max}=20$, $\gamma = 0.95$, $\delta = 0.01$, and $K=5$, the maximum number of different effects possible for any action.

\paragraph{Results} In \ref{fig: taxi-oo-results-appendix}, we show the cumulative reward curves for DOORmax agents in object-oriented Taxi. The RLang program, shown below, describes the full reward function and partial transition dynamics, specifically what happens when the agent tries to drive into walls and what happens when a passenger is picked up and dropped off. The plot shows a significant speedup for the RLang-enabled agent, even when partial transition dynamics are described.

\begin{lstlisting}[numbers=left]
Effect movement_effect:
    if S.taxi.touch_n and A == move_n:
        S' -> S
    if S.taxi.touch_s and A == move_s:
        S' -> S
    if S.taxi.touch_e and A == move_e:
        S' -> S
    if S.taxi.touch_w and A == move_w:
        S' -> S

Effect main:
    if S.taxi.on_passenger and A == pick_up:
        S'.passenger.in_taxi -> True
    if S.passenger.in_taxi and A == drop_off:
        S'.passenger.in_taxi -> False
        if S.taxi.on_destination:
            Reward 20
        else:
            Reward -10
    elif A == pick_up or A == drop_off:
        Reward -10
    else:
        Reward -1
\end{lstlisting}

\subsection{2D Minecraft} \label{appendix: minecraft}

2D Minecraft is a crafting environment based on \citeauthor{andreas2017modular} implemented as a $10\times 10$ grid. 
The state vector include a map of the environment, an inventory vector and the change on inventory with respect to the previous time step. The map is represented by $10 \times 10 \times 22$ tensor that represent with a one-hot vector the element at position $(x,y)$. 
The agent has $4$ actions to move in the cardinal directions by one position and a special action \texttt{use} to interact with the element \textit{in front}, i.e., given the current position and orientation of the agent, the position with which the agent interacts in the one the agent is facing. If such element is a primitive, the agent adds it to its inventory; if the element is a workbench and it has any of the primitive elements to build an object in the workbench, then those objects are built and added to the inventory (any primitive element used is removed from the inventory); if the agent is in front of water and has a bridge, it can \texttt{use} it and cross the water. If the agent has to interact with stone, it needs an axe in inventory. This is a goal-oriented environment; when the agent has the goal object in inventory, it receives a reward of $1$ and the episode terminates. When an episode starts, the agent is randomly placed in any free cell of the grid. We use a discount factor $\gamma=0.99$.

\paragraph{RLang-informed hierarchical DDQN Agent} To solve this environment, analogously to the Taxi environment, we use a hierarchical agent based on options.  We use DDQN \cite{van2016deep} as the algorithm to learn both the policy over options and intra-option policies. Algorithm \ref{alg:hrl-init} is used to initialized the intra-option policies. To implement DDQN and its hierarchical variation based on options, we based it on the Autonomous Learning Library \cite{nota2020autonomous}.

\paragraph{Neural Network architecture and parameters} We use a CNN with $4$ ReLU-activated layers with filter banks of size $32, 32, 32, 64$, a kernel size of $3$ and stride of $2$ (padding was used to keep the dimension $10\times 10$). The inventory and inventory change were processed with a ReLU MLP with hidden layer of size $32$ and output $32$. These two vectors are concatenated and passed through a linear layer of size $256$ (for the flat DDQN) and $64$ for the agents in the hierarchical DDQN implementation. Finally, this output vector is passed through a ReLU MLP with a hidden layer of size $64$ to get the value predictions for each action.

\paragraph{Hyperparameters} For DDQN, we use a linear schedule for $\epsilon-$greedy exploration with start with $\epsilon=1$ and $\epsilon=0$ and final exploration step $10000$. We use a learning rate $0.001$ and mini-batch of size $64$. We use a Prioritized replay buffer of size $10000$. The target network update frequency is $100$ steps. We set a time of $1000$ steps. 

For hierarchical DDQN: 
\begin{itemize}
    \item Outer Agent (policy over options): we use a linear schedule for $\epsilon-$greedy exploration with start with $\epsilon=1$ and $\epsilon=0$ and final exploration step $60000$. We use a learning rate $10^{-5}$ and mini-batch of size $64$. We use a Prioritized replay buffer of size $10000$. The target network update frequency is $100$ steps. This outer agent had a timeout of $1000$ steps;
    \item Inner Agents (intra-option policies): we use a linear schedule for $\epsilon-$greedy exploration with start with $\epsilon=1$ and $\epsilon=0.001$ and final exploration step $30000$. We use a learning rate $10^{-4}$ and mini-batch of size $128$. The target network update frequency is $100$ steps. We use a Prioritized replay buffer of size $10000$. Each subpolicy had a timeout of $100$ steps. 
\end{itemize}

\subsection{Classic Control}\label{appendix: classic-control}

\begin{figure*}[ht]
    \centering
    \begin{subfigure}[b]{0.4\textwidth}
         \centering
         \includegraphics[width=\textwidth]{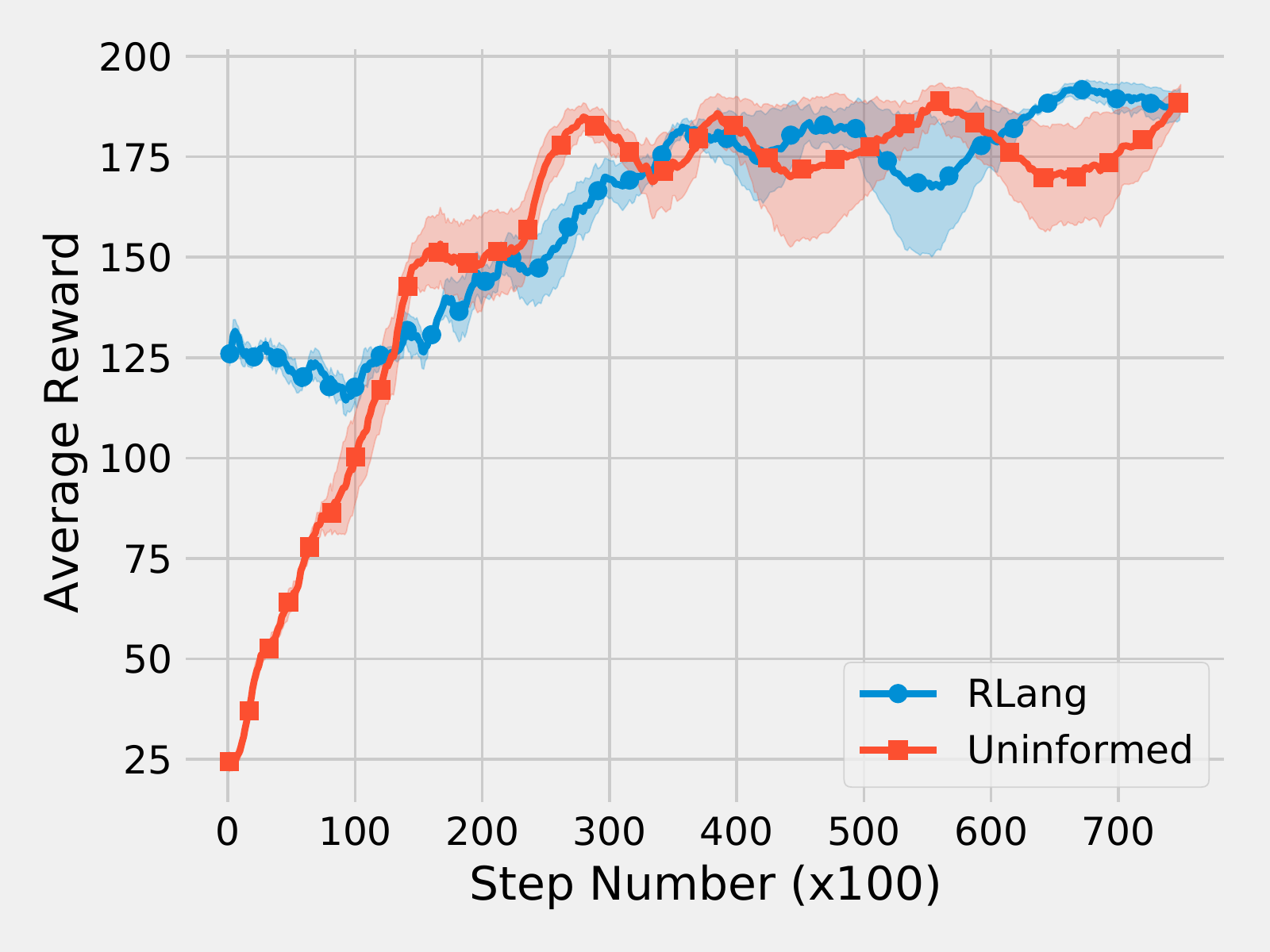}
         \caption{CartPole using REINFORCE}\label{fig: cartpole-results}
     \end{subfigure}
    \begin{subfigure}[b]{0.4\textwidth}
         \centering
         \includegraphics[width=\textwidth]{images/lunarlander/mean-reward.pdf}
         \caption{Lunar Lander using PPO}
     \end{subfigure}
    \caption{Average return curves for classic control tasks with continuous state spaces. We use RLang to provide the agent with an initial policy (non-optimal) that the agent can leverage to improve learning performance. We compare with the uninformed counterparts.} 
\end{figure*}

\paragraph{Environments} In this section, we consider Cartpole and Lunar Lander, two classic environments for RL research that have continuous state-space and discrete action space. For both environments, we use OpenAI Gym's implementations \cite{openaigym}, i.e. CartPole-v0 and LunarLander-v2.

Cartpole \cite{barto1983neuronlike} consists of an underactuated pole attached to a cart. At the beginning, the pole starts in a vertical position ($0$ degrees) and the agent has to learn a policy to keep it within $15$ degrees. The state consists of the position and the velocity of the cart, and the angle and angular velocity of the pole. The action space is to apply momentum to move the cart to the right or to the left. An episode ends when the pole absolute angle is greater than $15$ degrees, the position of the cart is greater than 2.4 units or after $200$ time steps. The agent gets a reward of $1$ every time step. This task is considered solved is the average return is at least $195$ over $100$ episodes. 

Lunar Lander simulates the task of landing a ship in a landing pad on the moon. The state consists of the ship's position and velocity, angle and angular velocity, and Boolean flags that indicate if the ship's leg is in contact with the ground. The actions are to fire the main engine, the right orientation engine, the left orientation engine and doing nothing. The agent gets a reward of $-100$ if the ship crashed and $+100$ if it lands correctly. It receives $+10$ for each leg that touches the ground and $-0.3$ if the main engine is fired. The task is solved with a return of at least $200$.

\paragraph{RLang-informed Policy Gradient Agent} To solve these environments, we provided non-optimal policies through RLang programs and use policy gradients methods to leverage this knowledge while learning. Algorithm \ref{alg:policy-mixing-alg} is derived from \cite{fernandez2006probabilistic}, in which, we probabilistically share control between the learning policy $\pi_\theta$ and the RLang-provided policy $\hat{\pi}$. At each time step, we choose which policy to follow by drawing a sample from a Bernoulli distribution with parameter $\beta$. We use such probabilistic mixing to collect trajectories and then optimize, $\theta$, using policy gradient methods. We use REINFORCE \cite{williams1992simple} for Cartpole and PPO \cite{schulman2017proximal} for Lunar Lander. The mixing parameter $\beta$ is annealed exponentially using a decay rate $\alpha$.

\begin{algorithm}[h]
\caption{Hierarchical Agent Initialization} \label{alg:policy-mixing-alg}
\begin{algorithmic}[1]
\FUNCTION {PolicyMixing(RLangKnowledge,decay\_rate)}

    \STATE Trajectories $\gets$ \texttt{Rollout}$(\text{Env}, \pi_\theta, \hat{\pi}, \beta)$
    \STATE $\theta \gets $\texttt{PolicyGradient}(Trajectories, $\pi_\theta$)
    \STATE $\beta \gets \beta * \text{decay\_rate} $
\ENDFUNCTION
\end{algorithmic}
\end{algorithm}

\paragraph{Neural Network architecture and parameters} For Cartpole and REINFORCE, we use a policy network using an MLP with hidden size $64$ and Leaky ReLU activations \cite{maas2013rectifier} with parameter $0.2$ 

In the case of Lunar Lander and PPO, we use a policy network based on a MLP with hidden size $64$ and Leaky ReLU activations \cite{maas2013rectifier} with parameter $0.2$. As a value network, we use an MLP with hidden size $64$ and Tanh activations.

\paragraph{Hyperparameters} For Cartpole, we use PFRL's REINFORCE implementation \cite{pfrl}. We use an initial mixing parameter $\beta=0.7$ and a decay rate $\alpha=0.99$. For REINFORCE, we use a learning rate of $0.001$ and a batch size of $5$. In Lunar Lander, we use PFRL's PPO implementation and use a mixing parameter $\beta=0.5$ and a decay rate $\alpha=0.9999$. For PPO, we used a learning rate of $0.0002$.

\paragraph{Cartpole Results} For Cartpole, we provide the RLang program below with a very simple prior policy. In Figure \ref{fig: cartpole-results}, we show the average return curves for RLang-informed REINFORCE and its uninformed performance, which show a jump-start performance gain for the agent then improves through experience.

\begin{lstlisting}[numbers=left]
    Policy balance_pole:
        if pole_angular_velocity > 0: 
            Execute move_right
        else:
            Execute move_left\end{lstlisting}


\end{document}